\PassOptionsToPackage{unicode}{hyperref}
\PassOptionsToPackage{hyphens}{url}
\documentclass[
]{article}
\usepackage[margin=1in]{geometry}
\usepackage{xcolor}
\usepackage{amsmath,amssymb}
\usepackage{iftex}
\ifPDFTeX
\usepackage[T1]{fontenc}
\usepackage[utf8]{inputenc}
\usepackage{textcomp} % provide euro and other symbols
\else % if luatex or xetex
\usepackage{unicode-math}
\fi
\usepackage{lmodern}
\makeatletter
\@ifundefined{KOMAClassName}{% if non-KOMA class
    \IfFileExists{parskip.sty}{%
        \usepackage{parskip}
    }{% else
        \setlength{\parindent}{0pt}
        \setlength{\parskip}{6pt plus 2pt minus 1pt}}
}{% if KOMA class
    \KOMAoptions{parskip=half}}
\makeatother
\usepackage{color}
\usepackage{fancyvrb}

\DefineVerbatimEnvironment{Highlighting}{Verbatim}{commandchars=\\\{\}}
\newenvironment{Shaded}{}{}

\newcommand{\CommentTok}[1]{\textcolor[rgb]{0.38,0.63,0.69}{\textit{#1}}}

\newcommand{\FloatTok}[1]{\textcolor[rgb]{0.25,0.63,0.44}{#1}}

\newcommand{\KeywordTok}[1]{\textcolor[rgb]{0.00,0.44,0.13}{\textbf{#1}}}
\newcommand{\NormalTok}[1]{#1}
\newcommand{\OperatorTok}[1]{\textcolor[rgb]{0.40,0.40,0.40}{#1}}

\newcommand{\VariableTok}[1]{\textcolor[rgb]{0.10,0.09,0.49}{#1}}

\usepackage{longtable,booktabs,array}
\usepackage{calc} % for calculating minipage widths
\usepackage{etoolbox}
\makeatletter
\patchcmd\longtable{\par}{\if@noskipsec\mbox{}\fi\par}{}{}
\makeatother
\IfFileExists{footnotehyper.sty}{\usepackage{footnotehyper}}{\usepackage{footnote}}
\makesavenoteenv{longtable}
\usepackage{graphicx}
\makeatletter
\newsavebox\pandoc@box
\newcommand*\pandocbounded[1]{% scales image to fit in text height/width
    \sbox\pandoc@box{#1}%
    \Gscale@div\@tempa{\textheight}{\dimexpr\ht\pandoc@box+\dp\pandoc@box\relax}%
    \Gscale@div\@tempb{\linewidth}{\wd\pandoc@box}%
    \ifdim\@tempb\p@<\@tempa\p@\let\@tempa\@tempb\fi% select the smaller of both
    \ifdim\@tempa\p@<\p@\scalebox{\@tempa}{\usebox\pandoc@box}%
    \else\usebox{\pandoc@box}%
    \fi%
}
\def\fps@figure{htbp}
\makeatother
\providecommand{\tightlist}{%
    \setlength{\itemsep}{0pt}\setlength{\parskip}{0pt}}
\ifPDFTeX\else
\fi
\IfFileExists{upquote.sty}{\usepackage{upquote}}{}
\IfFileExists{microtype.sty}{% use microtype if available
    \usepackage[]{microtype}
    \UseMicrotypeSet[protrusion]{basicmath} % disable protrusion for tt fonts
}{}
\usepackage{bookmark}
\IfFileExists{xurl.sty}{\usepackage{xurl}}{} % add URL line breaks if available
\makeatletter
\@ifundefined{xmpquote}{}{}
\makeatother
\usepackage{indentfirst}

\usepackage[most]{tcolorbox}

\hypersetup{
    hidelinks,
    pdfcreator={LaTeX via pandoc}}

\title{\textbf{VectorizationLLM: Smart Vectorization Based AI Assistant}}
\author{\textbf{R. Duke} \\ Department of Electrical \& Computer Engineering Technology\\New York Institute of Technology, Old Westbury, NY, USA \\ rduke01@nyit.edu }
\date{\today}

\begin{document}

    \maketitle

    \begin{abstract}
        VectorizationLLM is a specialized Large Language Model based on Google open-weight LLMs. The model is designed to assist students to learn smart vectorization, time/wave vector analysis,
        piecewise functions, Fourier analysis, and differential equations in MATLAB\cite{matlab}. The course application is CTEC 247: Applied Computational Analysis II by the Department of
        Electrical \& Computer Engineering Technology at New York Institute of Technology Old Westbury. The LLM model is designed to be an instructive assistant, providing detailed explanations
        of concepts with examples from in-class notes without providing direct answers to questions. The model is designed with a RAG (Retrieval Augmented Generation) knowledge base
        and system prompt architecture. Examples in both code, text, and images are provided in the LLM responses.
    \end{abstract}

    \newpage
    \tableofcontents
    \newpage

% =========================================================================
    \section{Introduction}
% =========================================================================
    \subsection{Purpose}
    ~\indent The course in question where MATLAB\cite{matlab} techniques for wave vectors, piecewise functions, Fourier analysis, and differential equations are applied is CTEC 247: Applied Computational Analysis II.
    This course is a junior-level course in the Electrical \& Computer Engineering Technology curriculum at New York Institute of Technology. The prerequisite course to CTEC 247 is
    CTEC 243: Applied Computational Analysis I, where vectorization, plotting, tables, and functions are taught. Students entering CTEC 247 spend one week in review of CTEC 243,
    followed by the first new unit of CTEC 247. The units in CTEC 247 are broken down into 6 numbered modules:

    \begin{itemize}
        \item \textbf{Module 0:} Review of CTEC 243
        \item \textbf{Module 1:} Boolean vectors and the find function
        \item \textbf{Module 2:} Piecewise Continuous Functions
        \item \textbf{Module 3:} Fourier Analysis
        \item \textbf{Module 4:} Solving of Symbolic Differential Equations (dsolve)
        \item \textbf{Module 5:} The ode45 solver of Differential Equations
    \end{itemize}

    Each module has its own major difficulty areas. The common difficulties in each unit year over year are listed below:

    \begin{itemize}
        \item \textbf{Module 0:} Recall of smart vectorization and indexing techniques
        \item \textbf{Module 1:} Using logical indexing instead of the find function
        \item \textbf{Module 2:} Not using the provided in-class toolbox functions, or concatenating functions incorrectly
        \item \textbf{Module 3:} Not using the provided in-class toolbox functions, not defining square waves properly
        \item \textbf{Module 4:} Confusing symbolic function syntax for time/wave vector syntax, calculating percentage overshoot/undershoot incorrectly
        \item \textbf{Module 5:} Not defining function handles properly, calculating percentage overshoot/undershoot incorrectly
    \end{itemize}

    ~\indent In the past, multiple methods were used to bridge the gap between these stated difficulties and instruction. In 2024, an
    instant messaging platform in Zoom\cite{zoom} Teams Chat was implemented to allow students to ask for help more
    efficiently. In 2025, additional examples were added to the main lectures. Neither of these methods were significantly effective to improve the
    bottom quarter of the class who were experiencing these difficulties. The methods only improved the high performing part of the class. Therefore,
    a more direct, user-friendly method of explaining concepts and providing examples is needed. An LLM bridges the gap between the Zoom\cite{zoom} chat method and
    the former examples method.

    \subsection{System Overview}
    ~\indent To provide a detailed overview of the VectorizationLLM system, two design aspects must be addressed. The first is the interface to the user, the second is the overview of the model
    itself. To provide the best interface to the user, the LLM is hosted using the OpenWebUi\cite{openwebui} platform. Each student is provided a user account tied to their university email.
    Since each student has their own account, the LLM provides a tailored, personalized experience for each prompt or question they ask the model. Furthermore, multiple prompts
    within the same thread build on each other. Therefore, if the student first asks about charting, then asks about piecewise continuous functions, the model will tie in the feedback
    from their initial charting prompt to their second prompt about piecewise continuous functions. Students are not allowed to upload files, as the model is intended to respond to MATLAB
    code pasted from their own scripts.

    ~\indent The VectorizationLLM model is based off of the Google DeepMind Gemma 4 26B A4B\cite{gemma} open weight LLM as a base, with a custom RAG and system prompt built into the interface. The recommended
    parameters for temperature=$1.0$, top\_p=$0.95$, and top\_k=$64$ are used to configure the base model. The RAG handles
    all technical data from the notes, as well as architectural data about the course (for example, what are the steps to performing Fourier analysis). The system prompt handles
    all interpretive data, including the instantiation of all numbered modules, image retrieval interface, among other features. Both these features are implemented using OpenWebUI's\cite{openwebui} default
    knowledge base and interface. An external RAG provider is not used in this implementation. However, all external images are hosted on a third party server.

    \section{Related Tools}
    ~\indent The contributions made by VectorizationLLM do not demonstrate scientific or architectural based novelty. Therefore, the contributions of this system exist within the realm of
    application based novelty. To evaluate tools with similar applications in engineering or education, three criteria must be evaluated. The first criteria is effective
    and accurate code writing, which involves the design and instruction of high level code concepts. The second criteria is detailed and accurate conceptual explanations, which
    include jargon and keywords relevant to related topics. The third criteria is academic integrity enforcement, which includes the omission of direct solutions to
    problems and examples suggested by the end user. All three criteria must be satisfied by VectorizationLLM or a related system to enable accurate and detailed
    assistance with course materials in a MATLAB curriculum. An effective AI assistant must be able to link by default the concepts with the code, while enforcing
    academic integrity. Several related tools exist which provide similar functionality, but fail at realizing one or more of these criteria.

    ~\indent The first related tool is also the most direct: the inline MATLAB\cite{matlab} AI code editor. As of MATLAB\cite{matlab} version 2025b,
    an advanced autofill AI code completion tool was introduced into the main code workspace. This tool predicts future lines of code
    based on previous code written in the scripts. Since this tool is primarily used for autofill, it satisfies the first and third criteria for code
    writing and academic integrity but fails at the second criteria for conceptual explanations. If a student writes an syntactically correct
    script but fails at the logic level of the course material, the MATLAB\cite{matlab} AI editor will not be able to explain
    why the code does not function correctly. Nevertheless, usage of the autofill tool is not discouraged as part of the course curriculum.

    ~\indent The next related toolset are the tools students would turn to first in the event of an issue: generalized LLMs. The usage of LLMs
    such as ChatGPT\cite{chatgpt} or Gemini\cite{gemini} are prohibited by policy in all fundamentals based programming courses, including all MATLAB courses.
    These tools fail at all three criteria unless properly prompted to do so. For example, generalized LLMs use loops or conditional statements in responses
    unless prompted to only use smart vectorization techniques. Furthermore, the key units in the course (Piecewise Continuous Functions, Fourier Analysis, dsolve)
    are not accurately searchable from standard LLM responses due to the models not having the correct functions or toolboxes as reference material. Therefore,
    a student would first have to make the LLM remember the course terminology before prompting for explanations. Lastly, generalized LLMs fail
    immediately on academic integrity criteria by always providing full solutions to problems. A student should not have to explain
    course topics to an AI assistant before asking for conceptual explanations.

    ~\indent The closest related toolset to VectorizationLLM is generalized or customizable AI tutors. AI tutors, such as CircleIn\cite{circlein}, ai-tutor.ai\cite{aitutor},
    or Khanmigo\cite{khanmigo} are designed as traditional course tutoring assistants. These tutors provide services such as conceptual recall, flash cards, or study guides.
    These tools are the closest in terms of functionality to VectorizationLLM, as they enforce academic integrity while prioritizing conceptual explanations, but their
    maximum scope is limited to specific materials and design architecture. CircleIn\cite{circlein} allows the upload of textbook materials and notes to the AI tutor, however
    its efficacy is undocumented. The existing demos for CircleIn\cite{circlein} show explanations of low-level mathematical concepts. They do not show the ability
    to link course concepts with high level code. In addition, CircleIn\cite{circlein} does not give the instructor the ability to fine-tune the AI assistant
    or system prompt. These design considerations make CircleIn's\cite{circlein} AI tutor appear to be for a more generalized approach to AI instruction. ai-tutor.ai\cite{aitutor} and
    or Khanmigo\cite{khanmigo} are more generalized than the CircleIn\cite{circlein} AI tutor. Both tools reference concepts primarily from the source company's knowledge base rather than
    course-specific materials. Furthermore, the level of customization provided by the ai-tutor.ai\cite{aitutor} and Khanmigo\cite{khanmigo} platforms is tied directly to
    the unique course material formats available. As a final note, all three AI tutors are hidden behind an individual or institutional paywall or subscription to the platforms,
    making implementation for only one course not optimal for an instructor.

    ~\indent The final related toolset to VectorizationLLM are external code generation tools, such as GitHub Copilot\cite{copilot} or Claude Code\cite{claude}. These
    tools immediately fail the second and third criteria for conceptual explanations and academic integrity, as they are mainly designed for full code generation. If a student
    does not understand the underlying concepts in the course material, using one of these tools can compound the student's logic errors creating incoherent solutions.
    Furthermore, all code should be generated in MATLAB\cite{matlab}, not on a GitHub\cite{copilot} repository. Lastly, Claude Code\cite{claude} requires
    a subscription to access full functionality. Similar to the drawback of the AI tutors, students should not have to pay to retrieve sufficient explanations for problems.

% =========================================================================
    \section{Architectural Goals \& Constraints}
% =========================================================================
    ~\indent The architectural design of the VectorizationLLM system focuses heavily on accommodating an entire class size while providing a tailored student experience.
    Therefore, the architectural goals and constraints resolve to the following:

    \begin{itemize}
        \item \textbf{Scalability:} Must support a class size of 20 students.
        \item \textbf{Interface:} Must be engaging and user-friendly to operate.
        \item \textbf{Accuracy:} Must provide accurate prompts while enforcing academic integrity.
    \end{itemize}

% =========================================================================
    \section{System Architecture}
% =========================================================================
    \subsection{RAG Architecture Diagram}
    The architecture follows a standard LLM RAG (Retrieval Augmented Generation) knowledge base and system prompt system design.

% Tip: Uncomment and link your architecture image below when using an editor
\begin{figure}[htbp]
     \centering
    \includegraphics[width=0.8\textwidth]{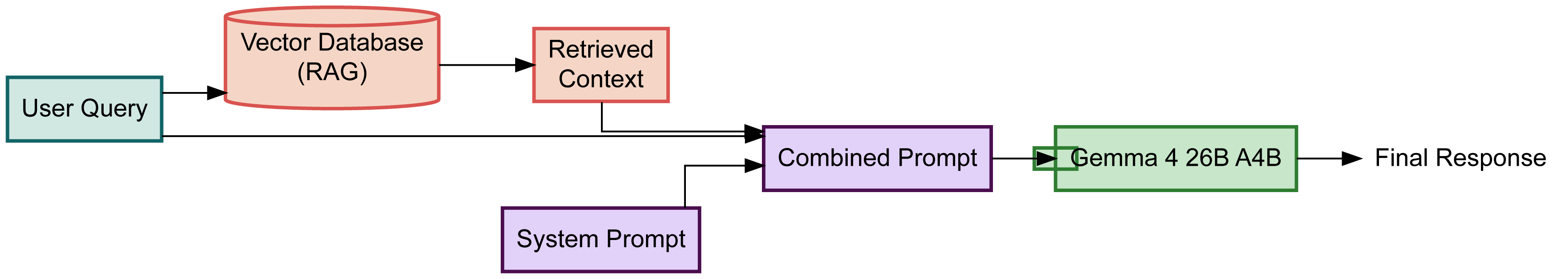}
    \caption{RAG Architecture}
    \label{fig:arch_diagram}
\end{figure}

    ~\indent Google Gemma 4 26B\cite{gemma} was chosen over other open weight LLM models like Qwen\cite{qwen} due to the high amount of user-friendliness
    which is expressed through prompt responses. Models like Qwen3.6 35B A3B\cite{qwen} provide more technical, stale responses which are less
    user friendly. Due to the implementation of the RAG and system prompt layers, the technical aspect of the model is preconfigured. Therefore, a model which is trained on
    highly technical data, but is less user friendly is less preferred over
    a user friendly model with slightly less technical training. To accommodate a class size of 20 students, OpenWebUI\cite{openwebui} was chosen as the web interface due to its intuitive RAG and user account control
    design. Furthermore, OpenWebUi\cite{openwebui} looks and feels like the interface from other Corporate LLMs like ChatGPT\cite{chatgpt} and Gemini\cite{gemini}. Therefore, the LLM interface
    provides a lower learning curve based on students' prior experiences.

    \newpage

    \subsection{System Prompt Design}
    ~\indent The system prompt design for VectorizationLLM is constructed with three main considerations: functionality, determinism, and guardrails. Each
    aspect forms a necessary framework to teach MATLAB\cite{matlab} code effectively.
    \begin{figure}[htbp]
        \centering
        \includegraphics[width=0.8\textwidth]{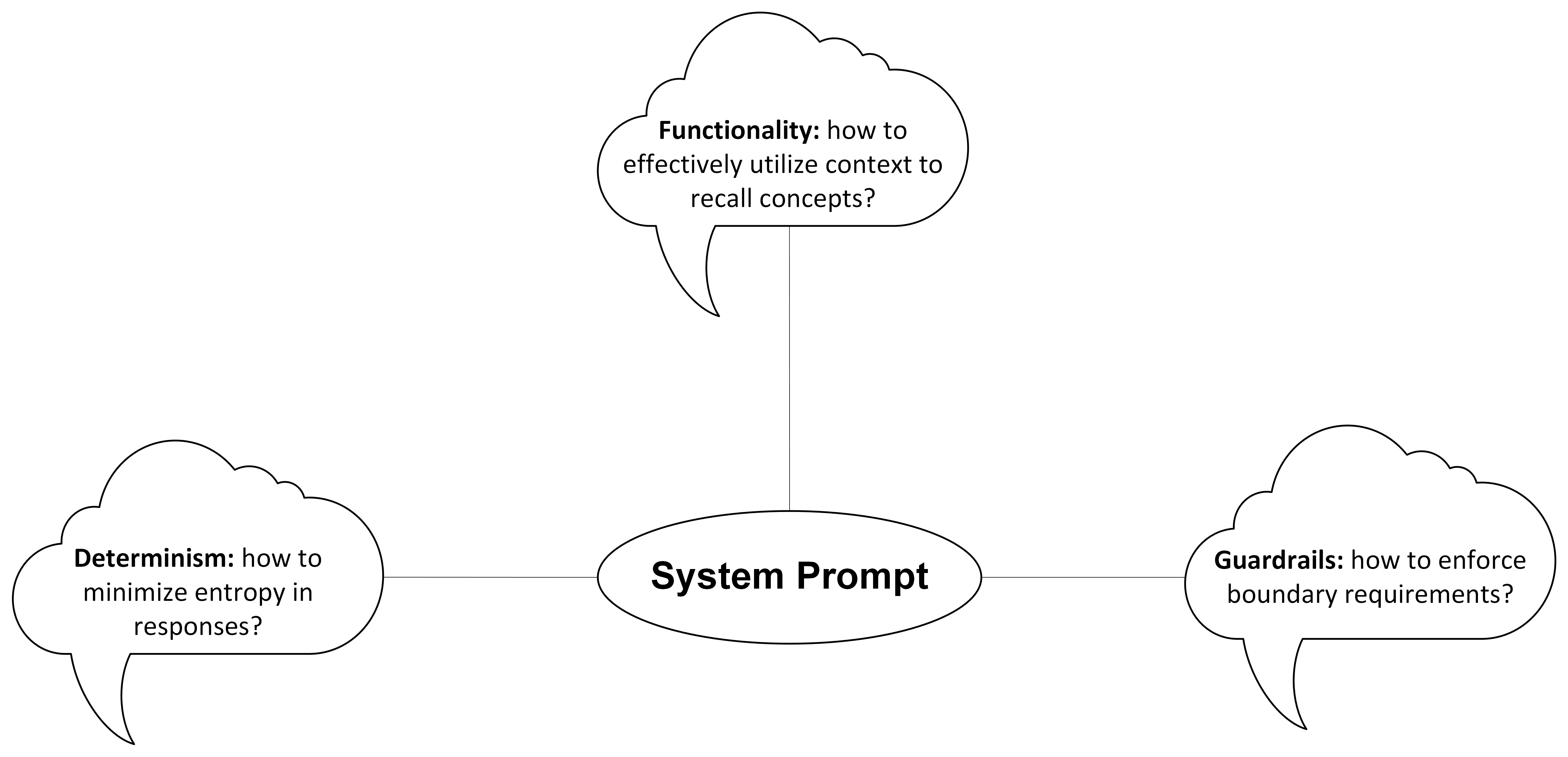}
        \caption{System Prompt Design}
        \label{fig:system_prompt}
    \end{figure}

    \paragraph{Functionality}~\\
    Functionality is designed via two main pillars: code explanations and code blocks.
    Code explanations are meant to be highly conceptual, often step-by-step explanations of MATLAB script and concept functionality. In addition, explanations of the interface
    (such as chart formatting and chart markers) are supplemented with example images.

    \begin{figure}[htbp]
        \centering
        \includegraphics[width=0.8\textwidth]{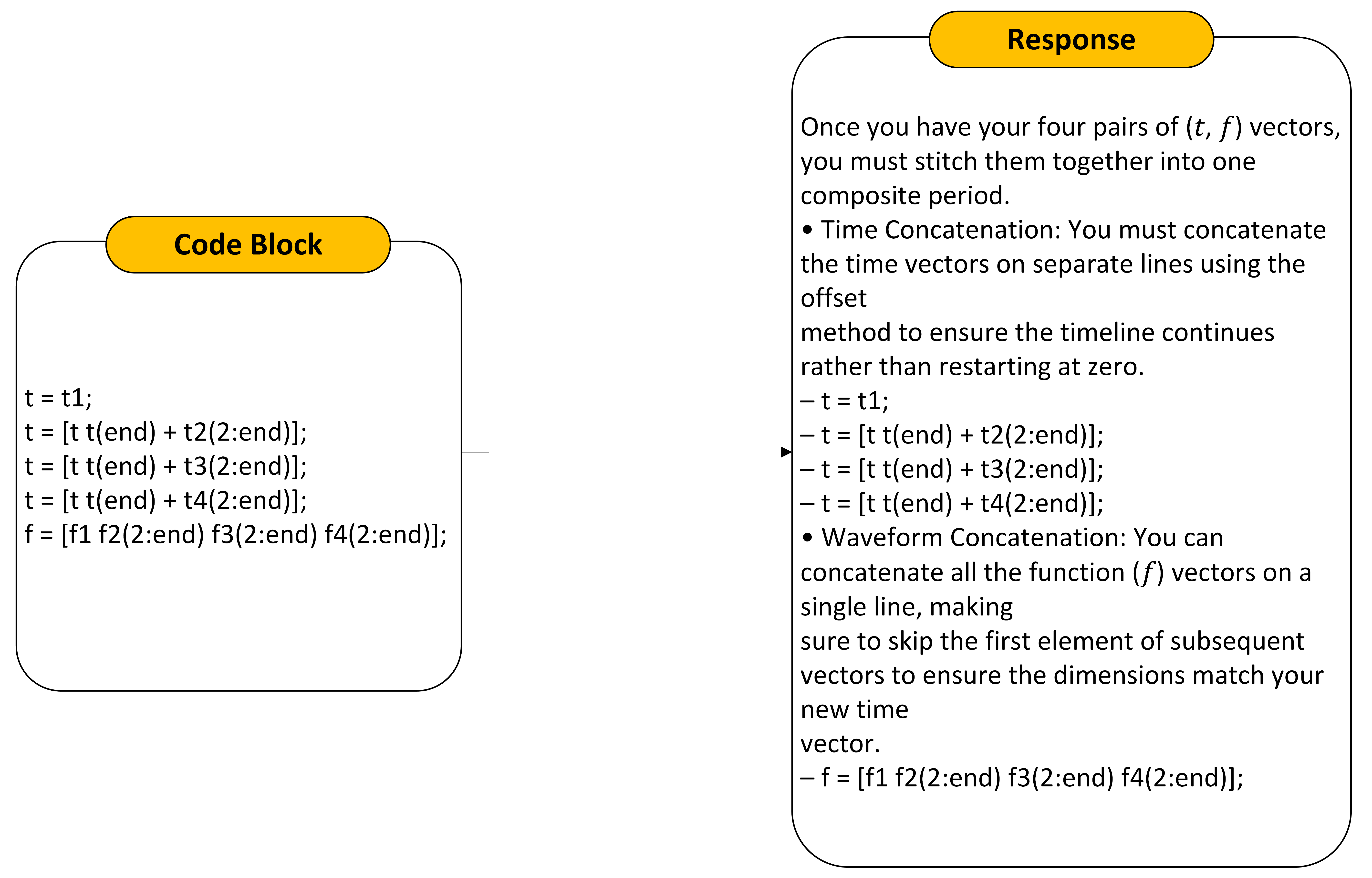}
        \caption{Sample Code Block with Response}
        \label{fig:code_block}
    \end{figure}

    ~\indent Code blocks are verbatim code snippets, either inline or in groups, taken from the class notes.
    Inline code snippets can be combined with code explanations to provide a deeper analysis. However, the system prompt design prioritizes code blocks over inline examples. Code
    explanations and code blocks can be generated from study-form requests to provide a roadmap for learning a specific topic or module. Study guides and roadmaps are intended
    to be an extra feature in which the system prompt is designed to recall topics based on topic name, alphanumeric identifier, or previous prompt context. This is a functionality
    that can benefit students if they did not get to ask a question during class, or if they failed to get the instructor to go back to a specific topic.
    \paragraph{Determinism}~\\
    ~\indent Determinism in VectorizationLLM is based on two main pillars: prompt accuracy and prompt history. Prompt accuracy is evaluated under two conditions: satisfactory
    responses based on minimum information (similar to a low effort, minimally descriptive student prompt), and how often satisfactory prompts appear when prompted with
    the same request. The system design requires precise tuning to recall specific information (code blocks, images, explanations) from the RAG material on the first prompt attempt,
    then retain that information across multiple future responses.
    \begin{figure}[htbp]
        \centering
        \includegraphics[width=0.8\textwidth]{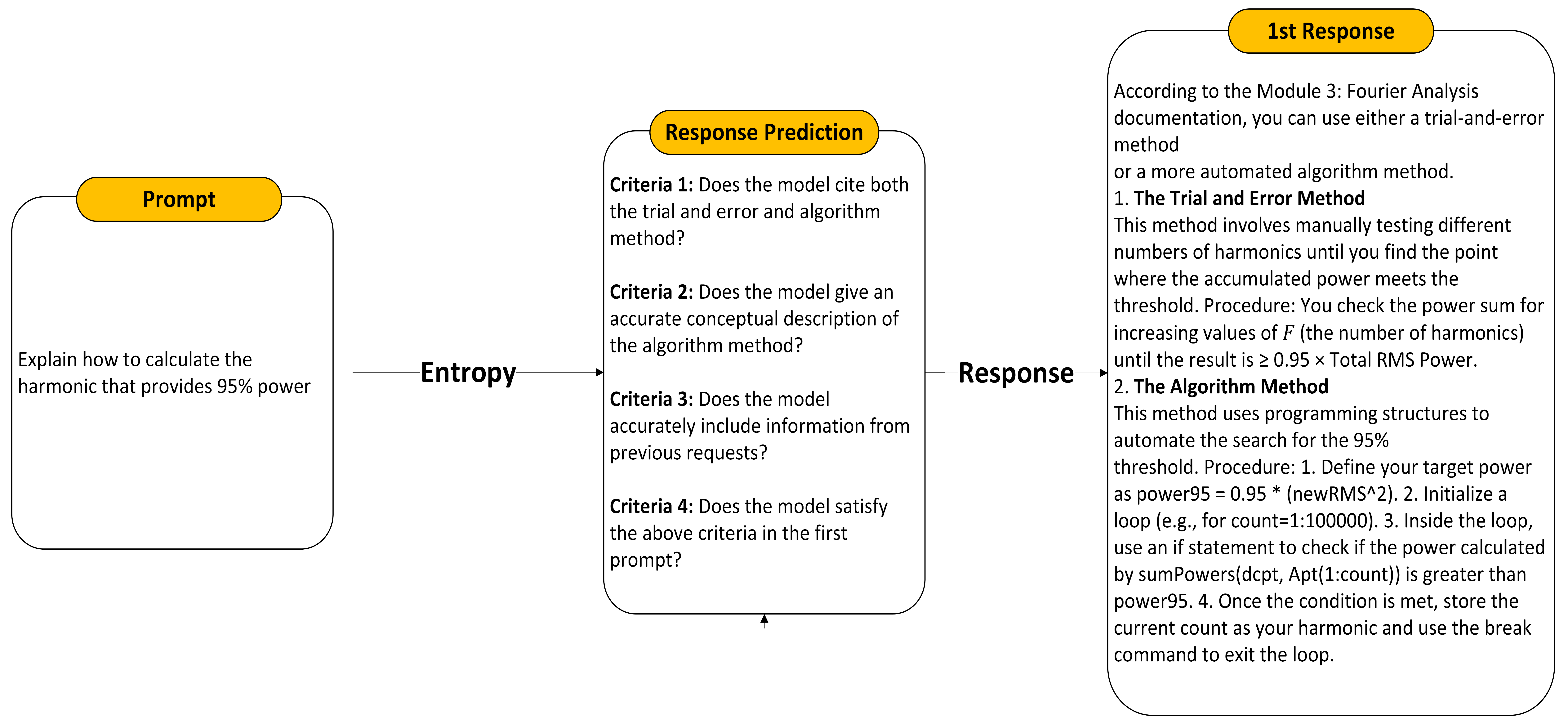}
        \caption{Logic Diagram for Deterministic Prompt Responses}
        \label{fig:determinism}
    \end{figure}

    ~\indent Due to the high verbosity of the class notes in the RAG material, the model
    requires full context mode in the RAG document retrieval system. Otherwise, the model will fail to pull the necessary data from the
    RAG material on the first attempt. Responses which state that the material does not exist, or responses that superimpose data from
    other modules can occur 2/3 of the time when full context mode is not enabled. Due to the high specificity of the system prompt design,
    failure cases in which data is invented by the model do not occur.

    ~\indent Cumulative prompts are possible
    (for example asking to explain step E in a module, where step E was identified in a previous response).
    The model clearly and accurately interprets prompt requests with minimal information (such as simply prompting \textit{Explain step E}). Furthermore, the LLM responds with
    non-technical comments which reference previous
    prompt requests made by the user in the same thread. Due to the nature of OpenWebUi\cite{openwebui}, separate threads are treated as separate contexts. Therefore,
    the model only recalls the history of the current thread. This is unlike corporate LLMs (Gemini\cite{gemini}, etc.), but recall capabilities across all threads is not required
    for this model.
    \paragraph{Guardrails}~\\
    ~\indent Guardrails in the system design must enforce course constraints, academic integrity, as well as provide LLM specific guardrails to the user. Course guardrails
    provide the basis for learning smart vectorization in MATLAB\cite{matlab}. The students must not be allowed to use loops or conditional statements. In addition,
    students must always use smart indexing over logical indexing, and use the toolbox functions
    over manually redefining them. There are always students at the bottom quarter of the class who attempt to use these forbidden techniques, so the system
    prompt is designed to strongly discourage the usage of such.
    \begin{figure}[htbp]
        \centering
        \includegraphics[width=0.9\textwidth]{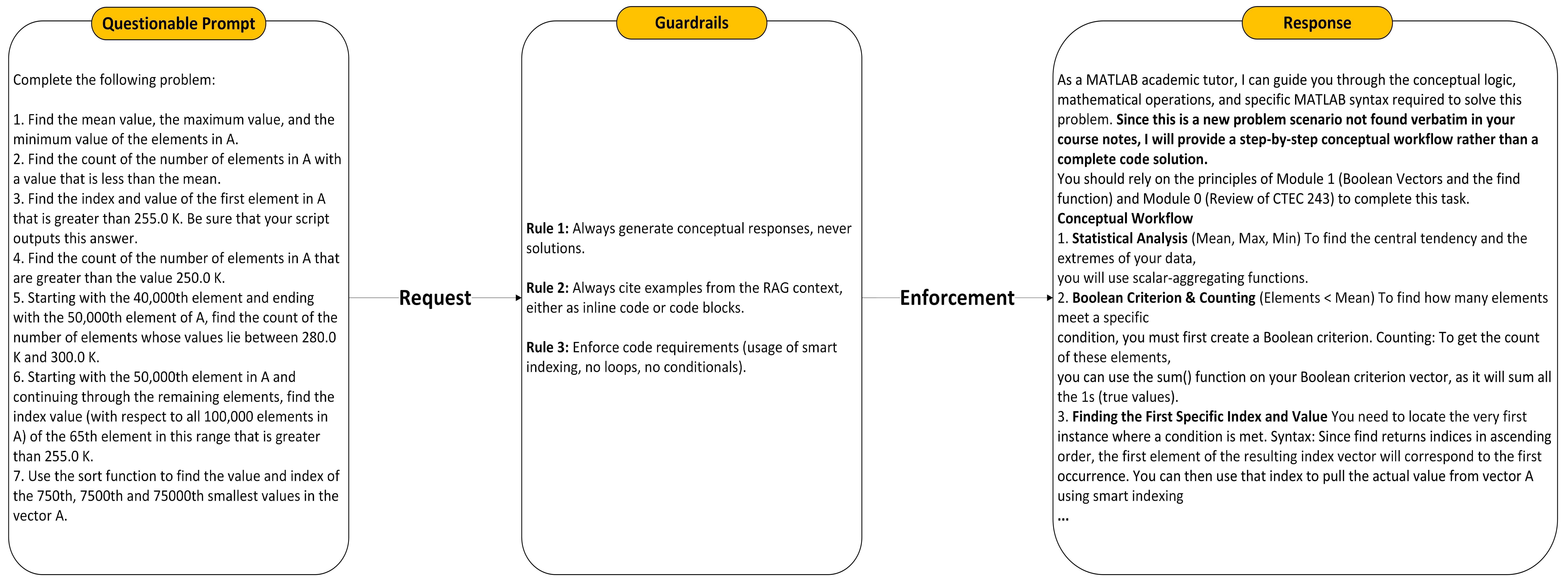}
        \caption{Logic Diagram for Guardrail Enforcement}
        \label{fig:guardrails}
    \end{figure}

    ~\indent Academic integrity
    guardrails provide necessary justification for the purpose of VectorizationLLM. Students must not be allowed to generate full solutions for code, nor generate new code from non-class examples.
    Any non-constrained LLM will violate this boundary immediately. The system prompt design is tailored to provide only code from the notes, and explains that it cannot generate
    solutions or new code when prompted with a homework or exam problem. Further LLM guardrails are implemented to prevent prompt jailbreaking/roleplay techniques. Therefore,
    the user cannot trick the model into revealing solutions. The now infamous Do Anything Now exploit\cite{doanything} is prevented in the system prompt design among other prompt
    jailbreaking techniques such as the user claiming that they are the professor.

    \subsection{RAG Documents}
    \begin{figure}[htbp]
        \centering
        \includegraphics[width=0.8\textwidth]{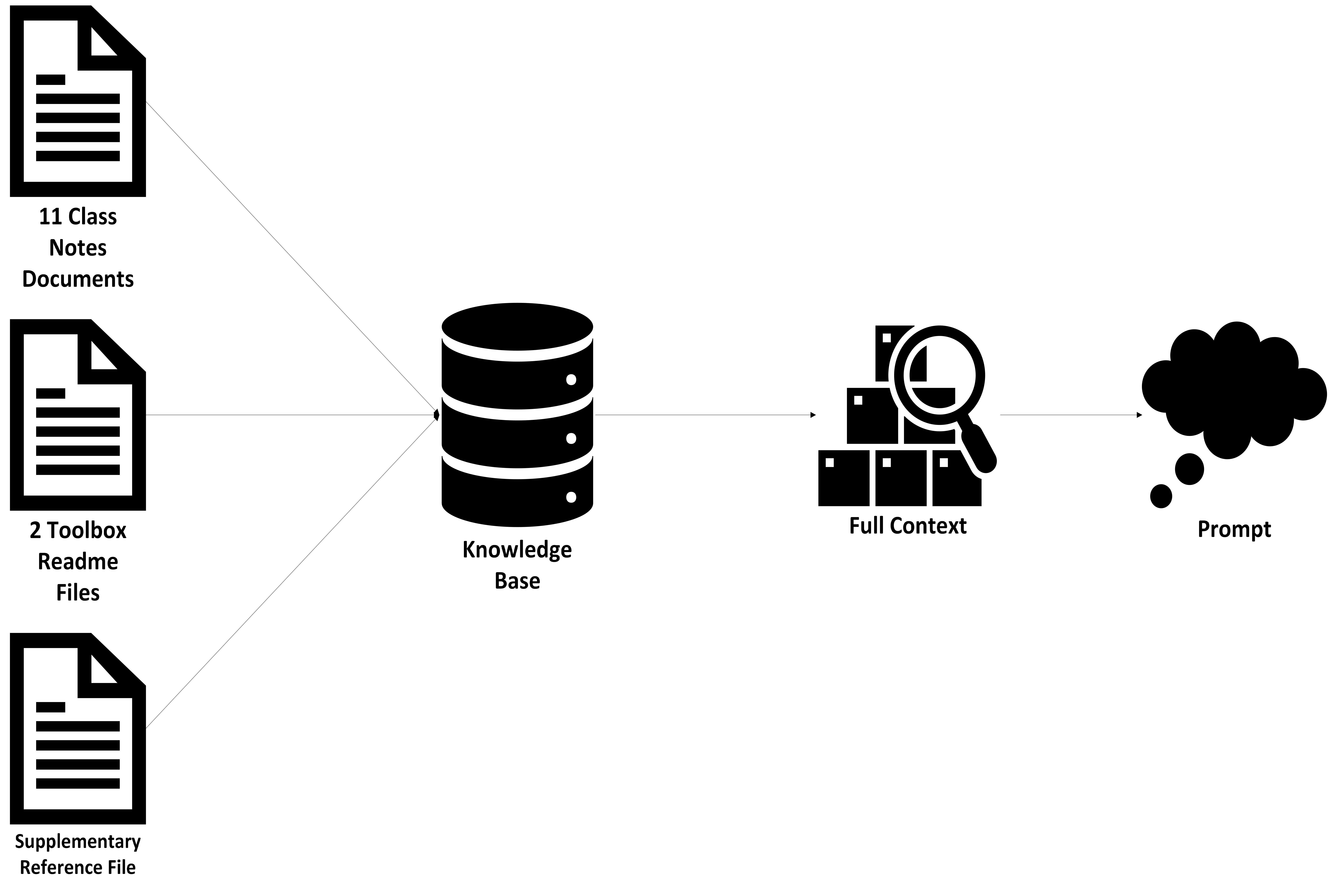}
        \caption{Logic Diagram for RAG Knowledge Base}
        \label{fig:rag}
    \end{figure}
    ~\indent The RAG documents input to the OpenWebUi\cite{openwebui} knowledge base consist of markdown versions of the class notes in both CTEC 243 and CTEC 247.
    All CTEC 243 prerequisite material is readily available to the model. For CTEC 247, all class notes and examples as well as a readme file
    for each of the toolboxes (piecewise continuous functions
    and Fourier analysis) are included in the context. There is one markdown file for each module in CTEC 243 and CTEC 247. The notes format consists of conceptual long form
    explanations paired with markdown code blocks as an example. The ratio of text explanations to code blocks is near 1:1. The examples follow
    case studies and procedures performed in class. In addition, supplementary documents which assist with RAG retrieval are included. For example, a document
    detailing how conversion divisors work (using a divisor on plot axes to scale them in SI units) is included to provide additional context for
    related problems.

    \subsection{AI Assistant Sample Prompts}
    The following sample prompts are answerable in VectorizationLLM:
    \begin{itemize}
        \item \textbf{Conceptual Explanation:} Explain the concatenation sequence in piecewise continuous functions.
        \item \textbf{Visualization:} Explain the calculation of percentage overshoot via chart markers.
        \item \textbf{Study Guide:} Explain how to learn Fourier Analysis in two weeks.
        \item \textbf{Extended Theory:} Explain how to calculate the voltage across the inductor coils with dsolve.
    \end{itemize}

    \subsection{Evaluation Metrics}
    ~\indent The LLM performance for code block recall was evaluated using a custom Python script. Since the system prompt and RAG framework prevent
    outside knowledge from entering the model responses, standard metrics based on hallucination (RAGAS\cite{ragas}, etc.) do not
    apply. The proper evaluation metrics for VectorizationLLM involve accurate concept and code block recall within the responses. Since
    the model must only use code blocks from the course notes, this evaluation script tests each code block in the responses and its relevance to
    verbatim or near verbatim recall from the notes. The responses tested in this methodology can be found in the appendix of this
    document.
    \begin{table}[htbp]
        \centering
        \begin{tabular}{cccl}
            \toprule
            \textbf{Block ID} & \textbf{Statement Matches} & \textbf{Verification Status} & \textbf{Code Block Snippet Sample} \\
            \midrule
            Block 01 & 3/3 & Pass & \texttt{criterion = wvLinear >= 2...} \\
            Block 02 & 3/3 & Pass & \texttt{criterion = tSpan > 10e-0...} \\
            Block 03 & 2/2 & Pass & \texttt{criterion = Vout == max(V...} \\
            Block 04 & 9/9 & Pass & \texttt{power95 = 0.95*(newRMS.\^{}2...} \\ % Fixed the caret symbol here
            Block 05 & 5/5 & Pass & \texttt{syms y(t)} \\
            Block 06 & 4/4 & Pass & \texttt{IC1 = y(0) == 5;} \\
            Block 07 & 1/1 & Pass & \texttt{fplot(q/C, [time\_start, t...} \\
            Block 08 & 2/2 & Pass & \texttt{vL = vP*sin(2*pi*f*t) - q...} \\
            Block 09 & 5/5 & Pass & \texttt{eqnI = L*Di + i*R == vP*s...} \\
            Block 10 & 1/1 & Pass & \texttt{syms i(t)} \\
            Block 11 & 6/7 & Fail & \texttt{A = 20; f = 2e03; T = 1/f...} \\
            Block 12 & 1/1 & Pass & \texttt{percentageOvershoot = ((3...} \\
            Block 13 & 1/1 & Pass & \texttt{percentageUndershoot = ((...} \\
            \text{Block 14} & 1/1 & Pass & \texttt{percentageUndershoot = ((...} \\
            Block 15 & 3/3 & Pass & \texttt{tSpan = 0:1e-06:70e-03;} \\
            \multicolumn{4}{|c|}{\textbf{Recall Rate:} 47/48 Code Lines in 15 Blocks or 97.92\%} \\
            \bottomrule
        \end{tabular}
        \caption{Detailed Verification Record for Code Blocks in Responses}
        \label{tab:rag_detailed_log}
    \end{table}

    ~\indent Based on the evaluation metrics, the recall rate for code lines is 47 out of 48 verbatim lines of code or 97.92\% in 15 total code blocks. The single outlier code
    block which recalled only 6/7 lines verbatim was based on the following response:

\newpage

\begin{tcolorbox}[colback=white, colframe=black, sharp corners,left=0cm, right=0cm]

    4. Example Workflow

    According to the \textbf{RLSinForce} example, the calculation follows
    this logic:

    \begin{Shaded}
        \begin{Highlighting}[]
            \CommentTok{\% Define constants}
            \VariableTok{A} \OperatorTok{=} \FloatTok{20}\OperatorTok{;} \VariableTok{f} \OperatorTok{=} \FloatTok{2e03}\OperatorTok{;} \VariableTok{T} \OperatorTok{=} \FloatTok{1}\OperatorTok{/}\VariableTok{f}\OperatorTok{;} \VariableTok{R} \OperatorTok{=} \FloatTok{500}\OperatorTok{;} \VariableTok{L} \OperatorTok{=} \FloatTok{39.8e{-}03}\OperatorTok{;} (Topic Reference Error)
            \VariableTok{syms} \VariableTok{i}\NormalTok{(}\VariableTok{t}\NormalTok{)}

            \CommentTok{\% Define the derivative and the differential equation}
            \VariableTok{Di} \OperatorTok{=} \VariableTok{diff}\NormalTok{(}\VariableTok{i}\OperatorTok{,} \FloatTok{1}\NormalTok{)}\OperatorTok{;}
            \VariableTok{vS} \OperatorTok{=} \VariableTok{A}\OperatorTok{*}\VariableTok{sin}\NormalTok{(}\FloatTok{2}\OperatorTok{*}\VariableTok{pi}\OperatorTok{*}\VariableTok{f}\OperatorTok{*}\VariableTok{t}\NormalTok{)}\OperatorTok{;}
            \VariableTok{eqnI} \OperatorTok{=} \VariableTok{L}\OperatorTok{*}\VariableTok{Di} \OperatorTok{+} \VariableTok{i}\OperatorTok{*}\VariableTok{R} \OperatorTok{==} \VariableTok{vS} \OperatorTok{{-}} \VariableTok{q}\OperatorTok{/}\VariableTok{C}\OperatorTok{;} \CommentTok{\% Note: q must be previously solved via dsolve}

            \CommentTok{\% Solve for current with initial condition i(0) = 0}
            \VariableTok{ICI} \OperatorTok{=} \VariableTok{i}\NormalTok{(}\FloatTok{0}\NormalTok{) }\OperatorTok{==} \FloatTok{0}\OperatorTok{;}
            \VariableTok{i} \OperatorTok{=} \VariableTok{dsolve}\NormalTok{(}\VariableTok{eqnI}\OperatorTok{,} \VariableTok{ICI}\NormalTok{)}\OperatorTok{;}
        \end{Highlighting}
    \end{Shaded}

\end{tcolorbox}

    In this response, the model provides a combined topic analysis from the ode45 unit and the dsolve unit. The model
    introduces constants for $A$, $f$, $R$, and $L$ which are from ode45 class notes, not the correct dsolve notes. This is
    the only response in which the model refers to two units at the same time inaccurately. However,
    it is important to note that this is a topic reference error, not a code reference error.
    The code and concepts shown are generated from the notes, not from the base LLM's pretrained knowledge.

    \section{Future Work}
    ~\indent Future work for VectorizationLLM depends on three main considerations: classroom adoption, implementation, and effectiveness. Student online time and
    session presence will be measured through OpenWebUi's\cite{openwebui} standard Admin Panel. Metrics will be recorded at the time of the midterm exam and the final exam. Student
    online time will be compared against the common difficulty points in each module, specifically students at the bottom quarter of the class. Student feedback
    will be monitored for any conceptual additions needed to the RAG material in future revisions of the model. The overall effectiveness of the model will be weighed
    against students' performance in previous years of CTEC 247.

    ~\indent After the Fall 2026 semester, CTEC 247 and its prerequisite CTEC 243 will be retired in the curriculum in favor
    of a unified MATLAB\cite{matlab} course: CTEC 249. A separate RAG knowledge base and system prompt has already been developed to allow the transition between courses to be seamless.
    The feedback and results from the LLM implementation in CTEC 247 will be considered in the future implementation of CTEC 249 in the Spring 2027 semester.

    \section{Source Material Disclosures}

    \subsection{Accessibility of Source Material}
    ~\indent To preserve academic integrity and protect the originality of the course material, the RAG source documents,
    system prompt design, and related image/source material used in VectorizationLLM are not available publicly.
    Access to the source material is limited to faculty/staff within the New York Institute of Technology College of Engineering \&
    Computing Sciences.

    \subsection{AI \& Machine Generated Content Disclosure}
    ~\indent As per the requirement for ethical publishing, the usage of AI and machine generated content in this manuscript
    must be disclosed. The only AI generated content are the VectorizationLLM responses in the Evaluation Metrics subsection
    and in the Appendix. No other content, including images, tables, or figures are generated by a machine. Furthermore, the system prompt and
    RAG knowledge base are based on an original design exclusive to the course material.

    \section{Conclusion}
    ~\indent VectorizationLLM is designed to be a course-specific academic assistant which supplements learning and understanding of high-level MATLAB\cite{matlab} concepts. The system prompt and RAG knowledge base
    are designed to prioritize and encourage proper course techniques and enforce academic integrity. The model provides detailed explanations of concepts while accurately citing
    relevant code blocks. The system is tested under minimally informative user prompts to ensure utility and effectiveness. The model accurately recalls
    and explains concepts within all units of CTEC 247. The first implementation of this model will occur in the Fall 2026 semester, which relevant
    student participation data and feedback will be collected to ensure future utility of the AI assistant.

    \section{Acknowledgments}
    ~\indent Acknowledgments go to the faculty of the Department of Electrical \& Computer Engineering Technology and the College of Engineering \& Computing
    Sciences at New York Institute of Technology.

    \newpage
    \bibliographystyle{IEEEtran}
    \bibliography{Bib}

\begin{thebibliography}{10}
\providecommand{\url}[1]{#1}
\csname url@samestyle\endcsname
\providecommand{\newblock}{\relax}
\providecommand{\bibinfo}[2]{#2}
\providecommand{\BIBentrySTDinterwordspacing}{\spaceskip=0pt\relax}
\providecommand{\BIBentryALTinterwordstretchfactor}{4}
\providecommand{\BIBentryALTinterwordspacing}{\spaceskip=\fontdimen2\font plus
\BIBentryALTinterwordstretchfactor\fontdimen3\font minus
  \fontdimen4\font\relax}
\providecommand{\BIBforeignlanguage}[2]{{%
\expandafter\ifx\csname l@#1\endcsname\relax
\typeout{** WARNING: IEEEtran.bst: No hyphenation pattern has been}%
\typeout{** loaded for the language `#1'. Using the pattern for}%
\typeout{** the default language instead.}%
\else
\language=\csname l@#1\endcsname
\fi
#2}}
\providecommand{\BIBdecl}{\relax}
\BIBdecl

\bibitem{matlab}
\BIBentryALTinterwordspacing
T.~M. Inc., ``Matlab version: 9.13.0 (r2022b),'' Natick, Massachusetts, United
  States, 2022. [Online]. Available: \url{https://www.mathworks.com}
\BIBentrySTDinterwordspacing

\bibitem{zoom}
\BIBentryALTinterwordspacing
{Zoom Video Communications, Inc.}, ``Zoom workplace,'' Computer software, 2026.
  [Online]. Available: \url{https://www.zoom.com}
\BIBentrySTDinterwordspacing

\bibitem{openwebui}
O.~W. Team, ``Open webui: An open, extensible, and usable interface for ai
  interaction,'' 2025.

\bibitem{gemma}
{Google DeepMind}, ``Gemma 4: Open models based on gemini research and
  technology,'' \url{https://ai.google.dev/gemma/docs/core/model_card_4}, 2026.

\bibitem{chatgpt}
\BIBentryALTinterwordspacing
{OpenAI}. (2026) Chatgpt (version 4.0). Large language model. Accessed: 23 June
  2026. [Online]. Available: \url{https://chat.openai.com}
\BIBentrySTDinterwordspacing

\bibitem{gemini}
\BIBentryALTinterwordspacing
Google, ``{Gemini 3.5 Flash},'' 2026, accessed: June 20, 2026. [Online].
  Available: \url{https://deepmind.google/models/gemini/flash/}
\BIBentrySTDinterwordspacing

\bibitem{circlein}
\BIBentryALTinterwordspacing
{CircleIn}. (2026) Circlein: The student success system. CircleIn. Mobile
  application and web study platform. [Online]. Available:
  \url{https://www.circleinapp.com/}
\BIBentrySTDinterwordspacing

\bibitem{aitutor}
\BIBentryALTinterwordspacing
{AI Tutor}, ``Ai tutor: Learn anything, anytime, anywhere,''
  \url{https://ai-tutor.ai/}, 2026, accessed: June 22, 2026. [Online].
  Available: \url{https://ai-tutor.ai/}
\BIBentrySTDinterwordspacing

\bibitem{khanmigo}
\BIBentryALTinterwordspacing
{Khan Academy}, ``Khanmigo: Ai-powered teaching assistant and tutor,'' 2023.
  [Online]. Available: \url{https://khanacademy.org}
\BIBentrySTDinterwordspacing

\bibitem{copilot}
{GitHub, Inc.}, ``Github copilot,'' \url{https://github.com}, 2026, computer
  software. Version X.Y.Z, accessed June 2026.

\bibitem{claude}
\BIBentryALTinterwordspacing
J.~Liu, X.~Zhao, X.~Shang, and Z.~Shen, ``Dive into claude code: The design
  space of today's and future ai agent systems,'' 2026. [Online]. Available:
  \url{https://arxiv.org/abs/2604.14228}
\BIBentrySTDinterwordspacing

\bibitem{qwen}
\BIBentryALTinterwordspacing
{Qwen Team}, ``{Qwen3.6-35B-A3B}: Agentic coding power, now open to all,''
  April 2026. [Online]. Available:
  \url{https://qwen.ai/blog?id=qwen3.6-35b-a3b}
\BIBentrySTDinterwordspacing

\bibitem{doanything}
\BIBentryALTinterwordspacing
X.~Shen, Z.~Chen, M.~Backes, Y.~Shen, and Y.~Zhang, ``"do anything now":
  Characterizing and evaluating in-the-wild jailbreak prompts on large language
  models,'' 2024. [Online]. Available: \url{https://arxiv.org/abs/2308.03825}
\BIBentrySTDinterwordspacing

\bibitem{ragas}
\BIBentryALTinterwordspacing
S.~Es, J.~James, L.~Espinosa-Anke, and S.~Schockaert, ``Ragas: Automated
  evaluation of retrieval augmented generation,'' 2025. [Online]. Available:
  \url{https://arxiv.org/abs/2309.15217}
\BIBentrySTDinterwordspacing

\end{thebibliography}
    \addcontentsline{toc}{section}{References}

    \newpage
    \section{Appendix: Sample Prompt History}
    \textbf{This sample prompt history was derived from a single thread with cumulative prompts, mimicking the personalized experience
    a student would have on the platform. The prompts in this set were designed as a minimum information problem, allowing the
    LLM to extrapolate the study responses based on minimal specificity. These prompts were executed from a user-level account on OpenWebUi\cite{openwebui}.}

    \subsection{Conversation 01: Study Guide: Course Topics}
    \subsubsection{USER}\label{user}

    What are the topics in this course?

    \subsubsection{ASSISTANT}\label{assistant}

    Based on the provided course documentation, CTEC 247 is organized into
    several modules covering a progression from MATLAB fundamentals to
    advanced signal analysis and circuit simulation.

    The topics included in this course are:

    \begin{itemize}
        \tightlist
        \item
        \textbf{Module 0: Review of CTEC 243}

        \begin{itemize}
            \tightlist
            \item
            Plotting techniques, including the use of conversion divisors for
            engineering units.
            \item
            Smart indexing and handling \texttt{t,\ wv} vector pairs.
            \item
            Functions and advanced plotting.
        \end{itemize}
        \item
        \textbf{Module 1: Boolean Vectors and Elementary Probability}

        \begin{itemize}
            \tightlist
            \item
            Introduction to Boolean expressions and the \texttt{find} function
            for criterion-based indexing.
            \item
            Statistical functions for advanced parsing (mean, median, mode,
            sort, min, and max).
        \end{itemize}
        \item
        \textbf{Module 2: Piecewise Continuous Functions}

        \begin{itemize}
            \tightlist
            \item
            Defining and concatenating piecewise waves using toolbox functions
            (like \texttt{makeRamp}, \texttt{makeSinewave}, \texttt{expRise},
            and \texttt{expDecay}).
            \item
            Creating complex waveforms and managing discontinuities.
            \item
            Using \texttt{makeCycles} to create periodic versions of piecewise
            continuous waves.
        \end{itemize}
        \item
        \textbf{Module 3: Fourier Analysis}

        \begin{itemize}
            \tightlist
            \item
            Generating square waves with specified duty cycles.
            \item
            Normalizing waveforms using RMS values and \texttt{get\_rms\_value}.
            \item
            Extracting Fourier data (frequencies, amplitudes, and phases) using
            the \texttt{fouriersA} toolbox function.
            \item
            Power analysis: calculating component power and proving the sum of
            harmonic powers matches the total RMS power.
            \item
            Approximating waveforms using \texttt{sumFouriers}.
            \item
            Determining the number of harmonics required to reach 95\% power.
        \end{itemize}
        \item
        \textbf{Module 4: MATLAB Symbolic Solution of CCLODEs}

        \begin{itemize}
            \tightlist
            \item
            Using \texttt{dsolve} to solve differential equations.
            \item
            Using \texttt{fplot} to obtain information from symbolic solutions.
            \item
            Transient response analysis, including calculating percentage
            overshoot and undershoot.
            \item
            RLC reactive circuit analysis (linking \(L\) with inductor terms,
            \(R\) with resistor terms, \(C\) with capacitor terms, and \(i\)
            with current terms).
        \end{itemize}
        \item
        \textbf{Module 5: MATLAB Function Handles \& the \texttt{ode45}
        Solver}

        \begin{itemize}
            \tightlist
            \item
            Introduction to function handles and the \texttt{@} operator.
            \item
            Using the \texttt{ode45} solver to solve RL and RC circuits with
            sinusoidal inputs.
            \item
            Using \texttt{ode45} to solve RLC circuits.
        \end{itemize}
    \end{itemize}

    \newpage
    \subsection{Conversation 02: Study Guide: CTEC 243 Review}
    \subsubsection{USER}\label{user-1}

    Explain what I have to review in CTEC 243.

    \subsubsection{ASSISTANT}\label{assistant-1}

    To prepare for CTEC 247, you should review the following fundamental
    MATLAB concepts from CTEC 243. These topics form the ``Module 0''
    foundation of the current course:

    \subsubsection{1. Plotting and Engineering
    Units}\label{plotting-and-engineering-units}

    You should be comfortable with the following plotting mechanics: *
    \textbf{Conversion Divisors:} Knowing how to scale axes or table columns
    to display values in engineering notation (e.g., dividing a time vector
    by \texttt{1e-03} to display in milliseconds). * \textbf{Basic Plotting
    Commands:} Using \texttt{plot(x,\ y)} with vector pairs, utilizing
    \texttt{hold\ on}/\texttt{hold\ off} to overlay multiple waveforms, and
    employing \texttt{axis}, \texttt{xlim}, and \texttt{ylim} to control
    chart limits. * \textbf{Chart Formatting:} Using \texttt{xlabel},
    \texttt{ylabel}, and \texttt{title} to label axes, and \texttt{grid\ on}
    to enable gridlines.

    \subsubsection{2. Indexing Techniques}\label{indexing-techniques}

    Proper indexing is critical for selecting specific data points or ranges
    within vectors: * \textbf{Smart Indexing:} Using single subscripts for
    scalars, ranges (e.g., \texttt{10001:30001}) for segments, and the
    \texttt{end} operator to dynamically access the last element of a
    vector. * \textbf{2D Matrix Indexing:} Understanding row and column
    subscripts \texttt{(row,\ col)} and how the colon operator \texttt{:}
    can select entire rows or columns. * \textbf{Reconstruction:} Using
    index vectors to reorder or extract specific elements from a vector to
    create new subsets.

    \subsubsection{3. Functions and
    Vectorization}\label{functions-and-vectorization}

    You must understand the difference between how MATLAB handles scalars
    versus vectors: * \textbf{Vector-Smart Functions:} Recognizing functions
    like \texttt{sin}, \texttt{exp}, \texttt{sqrt}, and \texttt{log} that
    apply an operation element-wise to every item in a vector. *
    \textbf{Scalar-Aggregating Functions:} Understanding that functions like
    \texttt{sum}, \texttt{prod}, and \texttt{length} collapse a vector into
    a single value. * \textbf{Element-wise Arithmetic:} Using the dot
    operator (\texttt{.*}, \texttt{./}, \texttt{.\^{}}) for element-wise
    multiplication, division, and powers to ensure mathematical operations
    are performed on individual vector components rather than performing
    matrix algebra. * \textbf{\texttt{t,\ wv} Vector Pairs:} Working with
    pairs of vectors where one represents time (\texttt{t}) and the other
    represents the waveform values (\texttt{wv}).

    \subsubsection{4. Data Organization and
    Tables}\label{data-organization-and-tables}

    \begin{itemize}
        \tightlist
        \item
        \textbf{Table Generation:} Using \texttt{fprintf} to create formatted,
        readable tables of instantaneous values by applying conversion
        divisors and specific field-width formatting.
        \item
        \textbf{The \texttt{ones} Function:} Using \texttt{K\ *\ ones(1,\ n)}
        to create vectors of constant values for use in tables or signal
        generation.
    \end{itemize}

    \newpage
    \subsection{Conversation 03: Conceptual Explanation: find Function}
    \subsubsection{USER}\label{user-2}

    Explain the syntax of the find function

    \subsubsection{ASSISTANT}\label{assistant-2}

    The \texttt{find} function in MATLAB is used to return the indices of
    elements in a matrix or vector that satisfy a specific logical
    criterion.

    Based on the provided documentation, here is the breakdown of its syntax
    and usage:

    \subsubsection{1. The Criterion}\label{the-criterion}

    Before using \texttt{find}, you must define a \textbf{criterion}, which
    must be a Boolean expression. A Boolean expression results in a vector
    of true/false (1/0) values. To create this criterion, you must directly
    compare a matrix or vector variable to a value or another expression
    using operators such as: * \textbf{Equality/Inequality:} \texttt{==}
    (equal to), \texttt{!=} (not equal to) * \textbf{Relational:}
    \texttt{\textgreater{}} (greater than), \texttt{\textless{}} (less
    than), \texttt{\textgreater{}=} (greater than or equal to),
    \texttt{\textless{}=} (less than or equal to) * \textbf{Logical:}
    \texttt{\&} (logical AND), \texttt{\textbar{}} (logical OR)

    \subsubsection{2. Function Syntax and
    Behavior}\label{function-syntax-and-behavior}

    The general workflow is: 1. \textbf{Define the criterion:}
    \texttt{criterion\ =\ variable\ \textgreater{}\ value;} 2. \textbf{Apply
    the find function:} \texttt{indexes\ =\ find(criterion);}

    \textbf{Key rules for the \texttt{find} function:} * \textbf{Returns
    Indices:} It returns the specific position (index) of every element that
    satisfies the criterion. * \textbf{Handling Empty Results:} If no
    elements in the matrix satisfy the criterion, the function returns an
    empty matrix. * \textbf{Direct Application:} These returned indices can
    be ``plugged back'' into the original matrix variable to extract the
    actual values, or into an equal-length matrix variable if you are
    working with a \texttt{t,\ wv} vector pair.

    \subsubsection{3. Examples of Usage}\label{examples-of-usage}

    \textbf{Range Selection} If you want to find all values in a vector
    \texttt{wvLinear} that fall between 2.5 and 3.5 inclusive, you would
    use:

    \begin{Shaded}
        \begin{Highlighting}[]
            \VariableTok{criterion} \OperatorTok{=} \VariableTok{wvLinear} \OperatorTok{\textgreater{}=} \FloatTok{2.5} \OperatorTok{\&} \VariableTok{wvLinear} \OperatorTok{\textless{}=} \FloatTok{3.5}\OperatorTok{;}
            \VariableTok{indexes} \OperatorTok{=} \VariableTok{find}\NormalTok{(}\VariableTok{criterion}\NormalTok{)}\OperatorTok{;}
            \VariableTok{values} \OperatorTok{=} \VariableTok{wvLinear}\NormalTok{(}\VariableTok{indexes}\NormalTok{)}\OperatorTok{;}
        \end{Highlighting}
    \end{Shaded}

    \textbf{Time Period Selection} To find all time periods in a
    \texttt{tSpan} vector between 10ms and 20ms (exclusive):

    \begin{Shaded}
        \begin{Highlighting}[]
            \VariableTok{criterion} \OperatorTok{=} \VariableTok{tSpan} \OperatorTok{\textgreater{}} \FloatTok{10e{-}03} \OperatorTok{\&} \VariableTok{tSpan} \OperatorTok{\textless{}} \FloatTok{20e{-}03}\OperatorTok{;}
            \VariableTok{indexes} \OperatorTok{=} \VariableTok{find}\NormalTok{(}\VariableTok{criterion}\NormalTok{)}\OperatorTok{;}
            \VariableTok{timePeriods} \OperatorTok{=} \VariableTok{tSpan}\NormalTok{(}\VariableTok{indexes}\NormalTok{)}\OperatorTok{;}
        \end{Highlighting}
    \end{Shaded}

    \textbf{Peak Detection} To find the indices where a waveform
    \texttt{Vout} is at its maximum or minimum value:

    \begin{Shaded}
        \begin{Highlighting}[]
            \VariableTok{criterion} \OperatorTok{=} \VariableTok{Vout} \OperatorTok{==} \VariableTok{max}\NormalTok{(}\VariableTok{Vout}\NormalTok{) }\OperatorTok{|} \VariableTok{Vout} \OperatorTok{==} \VariableTok{min}\NormalTok{(}\VariableTok{Vout}\NormalTok{)}\OperatorTok{;}
            \VariableTok{indexes} \OperatorTok{=} \VariableTok{find}\NormalTok{(}\VariableTok{criterion}\NormalTok{)}\OperatorTok{;}
        \end{Highlighting}
    \end{Shaded}

    \newpage
    \subsection{Conversation 04: Questionable Prompt: find Function Problem}
    \subsubsection{USER}\label{user-3}
    Complete the following problem:

    \textbf{Basic Instructions} Design a MATLAB script to generate the
    assigned outputs. Copy the output from the script execution and submit
    it separately in a Word document along with your script:
    \textbf{Preparation Work (downloads)} Download the file
    \textbf{temperatures.mat} from Canvas. This file is a randomized row
    vector of 100,000 temperatures in Kelvin from 247.039 K to 305.372 K.
    Then start the.m file for the assignment and put the contents of the
    row vector into your workspace. \textbf{Making it ``Word worthy''} Be
    sure to upload the latest output of your script as a word document to
    your submission, following the final changes you make. \textbf{Work to
    be Done} Write code to answer the questions that follow:
    \begin{enumerate}
        \def\labelenumi{\arabic{enumi}.}

        \item
      Find the mean value, the maximum value, and the minimum value of the elements in A.
        \item
      Find the count of the number of elements in A with a value that is less than the mean.
        \item
      Find the index and value of the first element in A that is greater than 255.0 K. Be sure that your script outputs this answer.
        \item
      Find the count of the number of elements in A that are greater than the value 250.0 K.
        \item
      Starting with the 40,000th element and ending with the 50,000th element of A, find the count of the number of elements whose values lie between 280.0 K and 300.0 K.
        \item
      Starting with the 50,000th element in A and continuing through the remaining elements, find the index value (with respect to all 100,000 elements in A) of the 65th element in this range that is greater than 255.0 K.
        \item
      Use the sort function to find the value and index of the 750th, 7500th and 75000th smallest values in the vector A.
    \end{enumerate}

    \subsubsection{ASSISTANT}\label{assistant-3}

    As a MATLAB academic tutor, I can guide you through the conceptual
    logic, mathematical operations, and specific MATLAB syntax required to
    solve this problem. Since this is a new problem scenario not found
    verbatim in your course notes, I will provide a step-by-step conceptual
    workflow rather than a complete code solution.

    You should rely on the principles of \textbf{Module 1 (Boolean Vectors
    and the \texttt{find} function)} and \textbf{Module 0 (Review of CTEC
    243)} to complete this task.

    \subsubsection{Conceptual Workflow}\label{conceptual-workflow}

    \textbf{1. Statistical Analysis (Mean, Max, Min)} To find the central
    tendency and the extremes of your data, you will use scalar-aggregating
    functions. * \textbf{Logic:} These functions take a vector and
    ``collapse'' it into a single representative value. *
    \textbf{Functions:} Use \texttt{mean()}, \texttt{max()}, and
    \texttt{min()} on your vector \texttt{A}.

    \textbf{2. Boolean Criterion \& Counting (Elements \textless{} Mean)} To
    find how many elements meet a specific condition, you must first create
    a Boolean criterion. * \textbf{Logic:} Create a logical vector where
    each element is \texttt{true} (1) if it is less than the mean and
    \texttt{false} (0) otherwise. * \textbf{Counting:} To get the
    \emph{count} of these elements, you can use the \texttt{sum()} function
    on your Boolean criterion vector, as it will sum all the \texttt{1}s
    (true values).

    \textbf{3. Finding the First Specific Index and Value} You need to
    locate the very first instance where a condition is met. *
    \textbf{Logic:} Create a criterion for values greater than 255.0 K. Use
    the \texttt{find} function to generate a vector of all indices that
    satisfy this. * \textbf{Syntax:} Since \texttt{find} returns indices in
    ascending order, the first element of the resulting index vector will
    correspond to the first occurrence. You can then use that index to pull
    the actual value from vector \texttt{A} using smart indexing.

    \textbf{4. Counting Elements \textgreater{} 250.0 K} This is similar to
    step 2. * \textbf{Logic:} Define a criterion (e.g.,
    \texttt{A\ \textgreater{}\ 250.0}). Apply the \texttt{sum()} function to
    this criterion to determine the total count of elements that satisfy the
    condition.

    \textbf{5. Range Selection and Nested Criteria} This requires combining
    smart indexing with a specific range. * \textbf{Logic:} First, extract
    the specific segment of the vector using smart indexing (from the
    40,000th to the 50,000th element). * \textbf{Logic:} Once you have this
    sub-vector, create a Boolean criterion to find values between 280.0 K
    and 300.0 K using the logical AND (\texttt{\&}) operator. Sum this
    criterion to find the count.

    \textbf{6. Advanced Indexing (Finding the Nth Occurrence in a Range)}
    This is a multi-step process involving range extraction and searching. *
    \textbf{Logic:} Extract the sub-vector starting from the 50,000th
    element to the end of \texttt{A}. * \textbf{Logic:} Within this
    sub-vector, create a criterion for values \textgreater{} 255.0 K and use
    \texttt{find} to get the indices \emph{relative to that sub-vector}. *
    \textbf{Logic:} Select the 65th index from that list. * \textbf{Logic:}
    To find the index with respect to the \emph{original} 100,000 elements,
    you must add the starting offset (the index of the first element of your
    sub-vector) to your result.

    \textbf{7. Sorting and Specific Rank Extraction} To find values at
    specific ``ranks'' (like the 750th smallest), you must reorder the data.
    * \textbf{Logic:} Use the \texttt{sort()} function to sort the vector
    \texttt{A} in ascending order. * \textbf{Syntax:} Note that
    \texttt{sort()} can return both the sorted values and their original
    indices. * \textbf{Logic:} Once sorted, you can use smart indexing to
    pick out the values and indices at the 750th, 7500th, and 75000th
    positions.

    \subsubsection{Summary of Required
    Functions/Operators:}\label{summary-of-required-functionsoperators}

    \begin{itemize}
        \tightlist
        \item
        \textbf{Functions:} \texttt{mean()}, \texttt{max()}, \texttt{min()},
        \texttt{sum()}, \texttt{find()}, \texttt{sort()}.
        \item
        \textbf{Operators:} \texttt{\textgreater{}} (greater than),
        \texttt{\textless{}} (less than), \texttt{\&} (logical AND).
        \item
        \textbf{Indexing:} \texttt{A(start:end)}, \texttt{A(index\_vector)}.
    \end{itemize}

    \newpage
    \subsection{Conversation 05: Study Guide: Piecewise Continuous Functions}
    \subsubsection{USER}\label{user-4}

    Explain how to learn piecewise continuous functions in two weeks

    \subsubsection{ASSISTANT}\label{assistant-4}

    Since I do not have a specific ``two-week syllabus'' in my
    documentation, I cannot provide a pre-made schedule. However, based on
    the \textbf{Module 2: Piecewise Continuous Functions} documentation, I
    can provide you with the high-level procedural workflow and the
    essential toolbox functions you must master to become proficient in this
    topic.

    To learn this subject effectively, you should focus on mastering the
    following conceptual steps and technical implementations.

    \subsubsection{1. Master the Core Toolbox
    Functions}\label{master-the-core-toolbox-functions}

    Before attempting to build complex waves, you must understand how to
    generate the fundamental ``building blocks'' using the piecewise
    toolbox. You should practice using: *
    \textbf{\texttt{makeRamp(A,\ B,\ tStart,\ tIncrement,\ tEnd)}}: Creates
    a linear wave between starting value \(A\) and ending value \(B\). *
    \textbf{\texttt{makeSinewave(amplitude,\ freq,\ phase,\ tStart,\ tIncrement,\ tEnd)}}:
    Generates a sinusoidal wave segment. *
    \textbf{\texttt{expRise(wvF,\ tau,\ incr,\ t\_end)}}: Creates a rising
    exponential wave. *
    \textbf{\texttt{expDecay(wv\_start,\ tau,\ incr,\ t\_end)}}: Creates a
    decaying exponential wave.

    \subsubsection{2. Learn the ``Single Period'' Construction
    Methodology}\label{learn-the-single-period-construction-methodology}

    Once you can create individual segments, you must learn how to stitch
    them together into a single piecewise period. The documentation outlines
    a specific 6-step process: * \textbf{Step A: Definition.} Define
    separate time and function vectors (\(t_1, f_1\) and \(t_2, f_2\), etc.)
    for each continuous segment. \textbf{Crucial Rule:} All segments must
    use the same time increment to ensure they can be concatenated
    correctly. * \textbf{Step B: Concatenation of Time.} Concatenate the
    time vectors. To avoid duplicating the point where segments meet, use
    the syntax: \texttt{t\ =\ {[}t1\ t1(end)\ +\ t2(2:end){]}}. *
    \textbf{Step C: Concatenation of Functions.} Concatenate the
    function/value vectors similarly: \texttt{f\ =\ {[}f1\ f2(2:end){]}}. *
    \textbf{Step D: Verification.} Plot the resulting \(t, f\) vector pair
    to ensure the segments connect as intended.

    \subsubsection{\texorpdfstring{3. Master Periodicity with
    \texttt{makeCycles}}{3. Master Periodicity with makeCycles}}\label{master-periodicity-with-makecycles}

    A piecewise function is often only the ``template'' for a repeating
    signal. You must learn how to take your single-period composite vector
    and turn it into a multi-cycle waveform. * \textbf{Function:} Use
    \texttt{{[}t,\ f{]}\ =\ makeCycles(t,\ f,\ periods)} to replicate your
    template a specified number of times.

    \subsubsection{4. Advanced Applications: Square Waves and
    Discontinuities}\label{advanced-applications-square-waves-and-discontinuities}

    To reach advanced proficiency, apply your knowledge to specific waveform
    types: * \textbf{Square Waves:} Practice creating square waves by
    combining \texttt{makeRamp} segments to control the \textbf{duty cycle}
    (the ratio of the high segment to the total period). * \textbf{Handling
    Discontinuities:} Learn how to use vertical lines or abrupt changes in
    \texttt{makeRamp} values to simulate instantaneous jumps in voltage or
    current.

    \textbf{Summary Study Checklist:} 1. Can I generate a single ramp or
    sine segment? 2. Can I concatenate two different segments without
    duplicating the joining point? 3. Can I create a repeating periodic wave
    using \texttt{makeCycles}? 4. Can I build a square wave with a specific
    duty cycle (e.g., 80\% or 50\%)?

    \newpage
    \subsection{Conversation 06: Conceptual Explanation: Waveform Concatenation}
    \subsubsection{USER}\label{user-5}

    Explain the concatenation sequence

    \subsubsection{ASSISTANT}\label{assistant-5}

    In MATLAB, the concatenation sequence is the method used to join
    individual vectors together to form a single, continuous composite
    vector. This is a critical process when working with \textbf{Piecewise
    Continuous Functions}, as it allows you to stitch together different
    functional segments (like a ramp and a sine wave) into one complete
    waveform.

    Based on the course documentation, the concatenation process is handled
    differently depending on whether you are joining time vectors or
    function (value) vectors.

    \subsubsection{\texorpdfstring{1. Concatenating Time Vectors
        (\(t\))}{1. Concatenating Time Vectors (t)}}\label{concatenating-time-vectors-t}

    When building a piecewise function, you cannot simply concatenate time
    vectors end-to-end using standard methods because the end of the first
    segment and the start of the second segment would create a duplicate
    point. To create a smooth transition, you must offset the subsequent
    time vectors.

    The standard sequence for time concatenation is: 1. Start with the first
    time vector: \texttt{t\ =\ t1;} 2. Append the next time vector, but skip
    its first element to avoid duplication and add the end value of the
    previous vector to shift its timeline:
    \texttt{t\ =\ {[}t\ t(end)\ +\ t2(2:end){]};}.

    This ensures that the time vector \(t\) continues chronologically from
    where the previous segment left off.

    \subsubsection{\texorpdfstring{2. Concatenating Function/Value Vectors
        (\(wv\))}{2. Concatenating Function/Value Vectors (wv)}}\label{concatenating-functionvalue-vectors-wv}

    Concatenating the waveform values (the \(wv\) or \(f\) vectors) is more
    straightforward because they are being joined to the already established
    time vector. You typically concatenate all segments in a single line.

    The standard sequence for value concatenation is:
    \texttt{f\ =\ {[}f1\ f2(2:end)\ f3(2:end){]};}.

    \textbf{Important Note on Dimensions:} It is vital that your \(t\) and
    \(wv\) vectors have the same dimensions. If your time concatenation
    results in a vector of a different length than your value concatenation,
    the pair will not work for plotting or further analysis.

    \subsubsection{3. Summary of the Single Period
    Methodology}\label{summary-of-the-single-period-methodology}

    When implementing a piecewise function for a single period (the ``6-Step
    Method''), the workflow follows this sequence : * \textbf{Step 1:} Set
    up the individual vectors for each segment (e.g., \(f_1(t_1)\),
    \(f_2(t_2)\)). * \textbf{Step 2:} (Optional) Plot each segment
    individually to verify the time axis is correct. * \textbf{Step 3:}
    Concatenate the time vectors using the offset method:
    \texttt{t\ =\ {[}t\ t(end)\ +\ t2(2:end){]};}. * \textbf{Step 4:}
    Concatenate the function vectors: \texttt{f\ =\ {[}f1\ f2(2:end){]};}. *
    \textbf{Step 5:} You now have a single \(t, f\) vector pair describing
    the piecewise function. * \textbf{Step 6:} Use \texttt{plot(t,\ f)} to
    visualize the result.

    \newpage
    \subsection{Conversation 07: Conceptual Explanation: Square Wave}
    \subsubsection{USER}\label{user-6}

    Explain how to create a square wave

    \subsubsection{ASSISTANT}\label{assistant-6}

    To create a square wave in MATLAB, you must utilize the techniques used
    for \textbf{Piecewise Continuous Functions}. A square wave is
    essentially a series of segments where the function value shifts between
    a high and a low state (or zero and a peak value) periodically.

    Based on the course documentation, the process involves three main
    conceptual stages:

    \subsubsection{1. Define the Segments (The ``Building
    Blocks'')}\label{define-the-segments-the-building-blocks}

    You first define the individual time and value segments that make up one
    single cycle of the square wave. A common way to do this is by using the
    \texttt{makeRamp} toolbox function to create the high and low portions
    of the wave.

    To control the \textbf{duty cycle} (the percentage of the period that
    the wave is ``high''), you must specify the duration of each segment: *
    \textbf{High Segment:} Create a ramp that stays at the peak value for a
    specific duration. * \textbf{Low Segment:} Create a ramp that stays at
    the base value for the remainder of the period. *
    \textbf{Transition/Vertical Line:} You can create a discontinuity (the
    vertical jump) by defining segments where the end value of one segment
    is different from the start value of the next.

    \textbf{Example of a 50\% Duty Cycle construction logic:} If your total
    period is \(T\): * The first segment (High) would run from \(0\) to
    \(0.25T\) at the peak voltage. * The second segment (Low) would run from
    \(0.25T\) to \(0.5T\) at zero voltage. * The third segment (Return to
    High) would run from \(0.5T\) back to \(0.75T\) at zero voltage, or use
    a third ramp to reset the cycle.

    \subsubsection{2. Concatenate into a Single
    Period}\label{concatenate-into-a-single-period}

    Once you have your individual \(t\) (time) and \(wv\) (waveform) vectors
    for each segment, you must stitch them together into one composite
    vector pair that describes one full cycle.

    \textbf{The Concatenation Sequence:} * \textbf{Time Vectors (\(t\)):}
    You cannot simply append them because you must avoid duplicating the
    timestamp where one segment ends and the next begins. You must use an
    offset: \texttt{t\ =\ {[}t1\ t1(end)\ +\ t2(2:end){]};}. * \textbf{Value
    Vectors (\(wv\)):} These are concatenated in a single line, also
    skipping the first element of subsequent vectors to maintain matching
    dimensions with the time vector: \texttt{wv\ =\ {[}wv1\ wv2(2:end){]};}.

    \textbf{Important Rule:} For this to work, all segments must have been
    created using the same time increment (step size).

    \subsubsection{3. Create Multiple Cycles}\label{create-multiple-cycles}

    The concatenation method described above only creates a single period.
    To create a continuous square wave for a longer duration, you must
    repeat that single period multiple times.

    \begin{itemize}
        \tightlist
        \item
        \textbf{Using \texttt{makeCycles}:} Once you have your single-period
        \(t, wv\) vector pair, you can use the \texttt{makeCycles} toolbox
        function to generate a specified number of periods.

        \begin{itemize}
            \tightlist
            \item
            \textbf{Input:} Your single-period \(t\) vector, your single-period
            \(wv\) vector, and the number of desired periods.
            \item
            \textbf{Output:} A new, much longer \(t, wv\) vector pair
            representing the periodic square wave.
        \end{itemize}
    \end{itemize}

    \subsubsection{Summary of Workflow}\label{summary-of-workflow}

    \begin{enumerate}
        \def\labelenumi{\arabic{enumi}.}
        \tightlist
        \item
        \textbf{Identify} the desired peak, shape, and duty cycle.
        \item
        \textbf{Generate} individual segments using toolbox functions like
        \texttt{makeRamp}.
        \item
        \textbf{Concatenate} the time vectors using the offset method and the
        value vectors by skipping the first element.
        \item
        \textbf{Verify} the single period by plotting with proper conversion
        divisors.
        \item
        \textbf{Expand} the waveform to multiple cycles using
        \texttt{makeCycles}.
    \end{enumerate}

    \newpage
    \subsection{Conversation 08: Questionable Prompt: Piecewise Wave Problem}
    \subsubsection{USER}\label{user-7}

    \begin{enumerate}
        \def\labelenumi{\alph{enumi})}
        \item
  Consider a base sinusoid waveform with amplitude A=20, period T=1ms, frequency 1KHz, and phase 0. Generate a piecewise decaying sinusoid with the following sinusoid components:
    \end{enumerate}

    f1(t) = A\emph{sin(2}pi\emph{freq}t1) for 0 \textless= t1 \textless= T
    f2(t) = 0.5\emph{A}sin(2\emph{pi}freq\emph{t2) for 0 \textless= t2
    \textless= T f3(t) = 0.25}A\emph{sin(2}pi\emph{freq}t3) for 0 \textless=
    t3 \textless= T f4(t) = 0.125\emph{A}sin(2\emph{pi}freq*t4) for 0
    \textless= t4 \textless= T

    Concatenate t1, t2, t3, t4 and f1, f2, f3, f4 in order to form the
    vector pair t, f.~Plot t, f with proper conversion divisors. Also plot 5
    cycles of t, f with proper conversion divisors. Format the plots with
    the following parameters:

    · Enable x and y gridlines. · X-axis Label: Time(ms) · Y-axis Label:
    Voltage(v) · Chart Title: Decaying Piecewise Sinusoid · Chart Title Font
    Size: 16pt

    \subsubsection{ASSISTANT}\label{assistant-7}

    To create this decaying piecewise sinusoid, you must follow the
    high-level procedural workflow for \textbf{Piecewise Continuous
    Functions}, specifically the \textbf{Single Period Method}.

    Since this is a new problem scenario, I will guide you through the
    conceptual steps, logic, and syntax required to achieve the desired
    output.

    \subsubsection{Step A: Definition of the Piecewise
    Waves}\label{step-a-definition-of-the-piecewise-waves}

    First, you must define the individual time and waveform vectors for each
    of the four segments.

    \begin{itemize}
        \tightlist
        \item
        \textbf{Important Rule:} All waves must have the same increment (step
        size) in their time vector to allow for proper concatenation.
        \item
        \textbf{Logic:} Use the \texttt{sinewave} function to generate each
        segment. Note that because each segment represents a new period in the
        piecewise sequence, you must define the time vectors for each segment
        starting from 0 to \(T\).
        \item
        \textbf{Parameters:}

        \begin{itemize}
            \tightlist
            \item
            Segment 1: Amplitude \(20\), Frequency \(1\text{kHz}\), Phase
            \(0^\circ\), Time \(0\) to \(T\).
            \item
            Segment 2: Amplitude \(10\) (which is \(0.5 \times 20\)), Frequency
            \(1\text{kHz}\), Phase \(0^\circ\), Time \(0\) to \(T\).
            \item
            Segment 3: Amplitude \(5\) (which is \(0.25 \times 20\)), Frequency
            \(1\text{kHz}\), Phase \(0^\circ\), Time \(0\) to \(T\).
            \item
            Segment 4: Amplitude \(2.5\) (which is \(0.125 \times 20\)),
            Frequency \(1\text{kHz}\), Phase \(0^\circ\), Time \(0\) to \(T\).
        \end{itemize}
    \end{itemize}

    You should use a range to define your time vectors (e.g.,
    \texttt{0:step:T}) rather than \texttt{linspace}.

    \subsubsection{Step B: Usage of the Concatenation
    Sequence}\label{step-b-usage-of-the-concatenation-sequence}

    Once you have your four pairs of \((t, f)\) vectors, you must stitch
    them together into one composite period.

    \begin{itemize}
        \tightlist
        \item
        \textbf{Time Concatenation:} You must concatenate the time vectors on
        separate lines using the offset method to ensure the timeline
        continues rather than restarting at zero.

        \begin{itemize}
            \tightlist
            \item
            \texttt{t\ =\ t1;}
            \item
            \texttt{t\ =\ {[}t\ t(end)\ +\ t2(2:end){]};}
            \item
            \texttt{t\ =\ {[}t\ t(end)\ +\ t3(2:end){]};}
            \item
            \texttt{t\ =\ {[}t\ t(end)\ +\ t4(2:end){]};}
        \end{itemize}
        \item
        \textbf{Waveform Concatenation:} You can concatenate all the function
        (\(f\)) vectors on a single line, making sure to skip the first
        element of subsequent vectors to ensure the dimensions match your new
        time vector.

        \begin{itemize}
            \tightlist
            \item
            \texttt{f\ =\ {[}f1\ f2(2:end)\ f3(2:end)\ f4(2:end){]};}
        \end{itemize}
    \end{itemize}

    \subsubsection{Step C: Plotting and Conversion
    Divisors}\label{step-c-plotting-and-conversion-divisors}

    To make the plot readable in engineering units, you must apply a
    conversion divisor to the time vector. * \textbf{Logic:} Since your time
    is in seconds and you want to display it in milliseconds (ms), divide
    the time vector by \texttt{1e-03}. * \textbf{Command:}
    \texttt{plot(t/1e-03,\ f);}

    \subsubsection{Step D: Creating Multiple
    Cycles}\label{step-d-creating-multiple-cycles}

    To create 5 cycles of this decaying pattern, use the \texttt{makeCycles}
    toolbox function. * \textbf{Syntax:}
    \texttt{{[}t5,\ f5{]}\ =\ makeCycles(t,\ f,\ 5);}

    \subsubsection{Step E: Chart Formatting}\label{step-e-chart-formatting}

    Finally, you must apply the specific formatting requested using MATLAB's
    plotting commands : * \textbf{Gridlines:} Use \texttt{grid\ on}. *
    \textbf{Labels:} Use
    \texttt{xlabel(\textquotesingle{}Time(ms)\textquotesingle{})} and
    \texttt{ylabel(\textquotesingle{}Voltage(v)\textquotesingle{})}. *
    \textbf{Title:} Use
    \texttt{title(\textquotesingle{}Decaying\ Piecewise\ Sinusoid\textquotesingle{},\ \textquotesingle{}FontSize\textquotesingle{},\ 16)}.

    \newpage
    \subsection{Conversation 09: Study Guide: Fourier Analysis}
    \subsubsection{USER}\label{user-8}

    Explain how to learn Fourier analysis in two weeks

    \subsubsection{ASSISTANT}\label{assistant-8}

    Since I do not have a pre-defined two-week syllabus, I cannot provide a
    specific calendar. However, based on the \textbf{Module 3: Fourier
    Analysis} documentation, I can provide a structured learning path
    organized by the high-level procedural steps required to master the
    subject.

    To learn Fourier analysis effectively, you should master these
    conceptual stages in order:

    \subsubsection{Week 1: Waveform Generation and
    Normalization}\label{week-1-waveform-generation-and-normalization}

    Before you can analyze the frequency components, you must be able to
    create accurate, normalized time-domain waveforms.

    \begin{itemize}
        \tightlist
        \item
        \textbf{Step 1: Master Piecewise Continuous Functions}

        \begin{itemize}
            \tightlist
            \item
            Learn to build square waves by defining segments with specific
            peaks, shapes, and duty cycles using toolbox functions like
            \texttt{makeRamp}.
            \item
            Practice the \textbf{Concatenation Sequence} to stitch segments
            together. Remember to use the offset method for time vectors
            (\texttt{t\ =\ {[}t\ t(end)\ +\ t2(2:end){]}}) to avoid duplicating
            time points.
        \end{itemize}
        \item
        \textbf{Step 2: Single-Cycle Plotting}

        \begin{itemize}
            \tightlist
            \item
            Practice plotting your single-period waveforms using proper
            \textbf{conversion divisors} (e.g., dividing by \texttt{1e-03} to
            display in milliseconds) to ensure the charts are readable.
        \end{itemize}
        \item
        \textbf{Step 3: RMS Normalization}

        \begin{itemize}
            \tightlist
            \item
            Learn to normalize your waveforms to a specific RMS voltage. You
            will need to use the \texttt{get\_rms\_value} toolbox function to
            find the current RMS and calculate a normalization factor.
            \item
            \textbf{Tip:} Always confirm your normalization by creating a table
            that displays the new amplitude and RMS value.
        \end{itemize}
    \end{itemize}

    \subsubsection{Week 2: Frequency Domain Extraction and Power
    Analysis}\label{week-2-frequency-domain-extraction-and-power-analysis}

    Once you can generate signals, you move into extracting and interpreting
    their spectral data.

    \begin{itemize}
        \tightlist
        \item
        \textbf{Step 4: Extracting Fourier Data}

        \begin{itemize}
            \tightlist
            \item
            Master the \texttt{fouriersA} toolbox function. This function
            provides the fundamental frequency, the \(N/2\) accessible
            harmonics, their amplitudes (\texttt{A}), and their phase angles
            (\texttt{theta}).
        \end{itemize}
        \item
        \textbf{Step 5: Power Component Analysis}

        \begin{itemize}
            \tightlist
            \item
            Learn to prove that the sum of the waveform's power components
            equals the total RMS power. In this course, power is calculated
            across a 1 Ohm resistor, so \(Power = V_{rms}^2/1\).
            \item
            Use the \texttt{sumPowers} toolbox function to facilitate these
            calculations.
        \end{itemize}
        \item
        \textbf{Step 6: Visualizing the Spectrum}

        \begin{itemize}
            \tightlist
            \item
            Practice creating \textbf{stem plots} of the power components and
            frequencies for a specified number of harmonics.
        \end{itemize}
        \item
        \textbf{Step 7: Waveform Approximation and 95\% Power}

        \begin{itemize}
            \tightlist
            \item
            Learn to reconstruct the waveform using the \texttt{sumFouriers}
            toolbox function, which takes the Fourier components and returns a
            \(t, wv\) vector pair of the approximated waveform.
            \item
            \textbf{The 95\% Power Goal:} Determine how many harmonics are
            required to reach 95\% of the total power. You can do this via two
            methods:

            \begin{enumerate}
                \def\labelenumi{\arabic{enumi}.}
                \tightlist
                \item
                \textbf{Trial and Error:} Manually summing harmonics until you
                reach the threshold.
                \item
                \textbf{The Algorithm Method:} Using a loop and conditional
                statements to find the specific harmonic where the power exceeds
                the 95\% threshold.
            \end{enumerate}
        \end{itemize}
        \item
        \textbf{Step 8: Phase and Shift Analysis}

        \begin{itemize}
            \tightlist
            \item
            Understand how time-shifting a waveform affects its components. A
            time shift will leave the power spectrum (amplitudes) unchanged but
            will cause the phase components (angles) to change.
        \end{itemize}
    \end{itemize}

    \newpage
    \subsection{Conversation 10: Conceptual Explanation: sumFouriers Function}
    \subsubsection{USER}\label{user-9}

    Explain the sumFouriers function

    \subsubsection{ASSISTANT}\label{assistant-9}

    The \texttt{sumFouriers} function is a toolbox utility used in Fourier
    analysis to reconstruct an approximated waveform by summing a specific
    number of its harmonic components.

    Based on the documentation, here is a detailed breakdown of the
    function:

    \subsubsection{\texorpdfstring{\textbf{Core
    Purpose}}{Core Purpose}}\label{core-purpose}

    The function takes the Fourier data (DC component, frequencies,
    amplitudes, and phases) and performs a summation to generate a
    time-domain waveform. It specifically returns a \(t, wv\) vector pair
    that represents the approximated version of the original signal.

    \subsubsection{\texorpdfstring{\textbf{Mathematical
    Model}}{Mathematical Model}}\label{mathematical-model}

    The function approximates the waveform using the following mathematical
    form for each harmonic:
    \[A(n) \cdot \cos(2\pi \cdot n \cdot f(n) \cdot t + \theta(n))\] Where:
    * \(A(n)\) is the amplitude of the \(n^{th}\) harmonic. * \(f(n)\) is
    the frequency of the \(n^{th}\) harmonic. * \(\theta(n)\) is the phase
    angle of the \(n^{th}\) harmonic.

    \subsubsection{\texorpdfstring{\textbf{Inputs and
    Outputs}}{Inputs and Outputs}}\label{inputs-and-outputs}

    \textbf{Inputs:} * \textbf{\texttt{N}}: The number of samples taken over
    a single period of the original signal. * \textbf{\texttt{dc}}: The
    scalar DC component (the offset) of the signal. * \textbf{\texttt{f}}: A
    vector of frequencies where \texttt{f(1)} represents the fundamental
    frequency. * \textbf{\texttt{A}}: A vector of magnitudes (amplitudes)
    corresponding to each harmonic. * \textbf{\texttt{theta}}: A vector of
    phase angles corresponding to each harmonic. * \textbf{\texttt{F}}: The
    specific number of harmonics (starting from the fundamental) that you
    wish to sum.

    \textbf{Outputs:} * \textbf{\texttt{t}}: A time vector representing one
    period of the approximated waveform. * \textbf{\texttt{w}}: The
    resulting waveform vector (the sum of the DC component and the specified
    \(F\) harmonics).

    \subsubsection{\texorpdfstring{\textbf{Internal Logic and
    Constraints}}{Internal Logic and Constraints}}\label{internal-logic-and-constraints}

    \begin{enumerate}
        \def\labelenumi{\arabic{enumi}.}
        \tightlist
        \item
        \textbf{Time Vector Generation:} The function calculates the period
        \(T\) from the fundamental frequency (\(T = 1/f(1)\)) and determines a
        time step (\(dt = T/N\)) to create the time vector \texttt{t}.
        \item
        \textbf{Initialization:} The waveform vector \texttt{w} is initialized
        as a vector of constant values equal to the \texttt{dc} component.
        \item
        \textbf{Safety Check:} The function includes a conditional check to
        ensure that the requested number of harmonics (\texttt{F}) does not
        exceed the maximum available harmonics (\(N/2 - 1\)).
        \item
        \textbf{Summation Loop:} It uses a loop to iterate from 1 to \(F\),
        adding the contribution of each successive harmonic to the \texttt{w}
        vector. \emph{(Note: Per course directives, while loops are generally
        discouraged, they are explicitly allowed for this specific algorithm
        in Fourier analysis).}
    \end{enumerate}

    \newpage
    \subsection{Conversation 11: Conceptual Explanation: 95\% Power Harmonic}
    \subsubsection{USER}\label{user-10}

    Explain how to calculate the harmonic that provides 95\% power

    \subsubsection{ASSISTANT}\label{assistant-10}

    There are two primary ways to determine at which harmonic a waveform
    reaches 95\% of its total power. According to the \textbf{Module 3:
    Fourier Analysis} documentation, you can use either a trial-and-error
    method or a more automated algorithm method.

    \subsubsection{1. The Trial and Error
    Method}\label{the-trial-and-error-method}

    This method involves manually testing different numbers of harmonics
    until you find the point where the accumulated power meets the
    threshold. * \textbf{Logic:} You use the \texttt{sumPowers} toolbox
    function to calculate the power of the DC component and the accumulated
    power of the harmonics. * \textbf{Procedure:} You check the power sum
    for increasing values of \(F\) (the number of harmonics) until the
    result is \(\geq 0.95 \times \text{Total RMS Power}\). * \textbf{Note:}
    This method is described as being more precise but requiring more work.

    \subsubsection{2. The Algorithm Method}\label{the-algorithm-method}

    This method uses programming structures to automate the search for the
    95\% threshold. * \textbf{Logic:} You utilize a \texttt{for} loop and a
    conditional \texttt{if} statement to iterate through the harmonics and
    stop once the criteria are met. * \textbf{Procedure:} 1. Define your
    target power as \texttt{power95\ =\ 0.95\ *\ (newRMS\^{}2)}. 2.
    Initialize a loop (e.g., \texttt{for\ count=1:100000}). 3. Inside the
    loop, use an \texttt{if} statement to check if the power calculated by
    \texttt{sumPowers(dcpt,\ Apt(1:count))} is greater than
    \texttt{power95}. 4. Once the condition is met, store the current
    \texttt{count} as your harmonic and use the \texttt{break} command to
    exit the loop. * \textbf{Note:} This method is described as being less
    precise but requiring less work.

    \subsubsection{Summary of Required Toolbox
    Functions}\label{summary-of-required-toolbox-functions}

    To perform either method, you will need the following functions from the
    Fourier toolbox: * \textbf{\texttt{sumPowers(dc,\ A)}}: Calculates the
    total power including the DC component and the amplitudes in vector
    \texttt{A}. * \textbf{\texttt{fouriersA(x,\ T)}}: Used initially to
    extract the necessary DC, frequency, amplitude, and phase data from your
    waveform.

    \newpage
    \subsection{Conversation 12: Conceptual Explanation: 95\% Power Harmonic Algorithm}
    \subsubsection{USER}\label{user-11}

    Explain the algorithm method

    \subsubsection{ASSISTANT}\label{assistant-11}

    The \textbf{algorithm method} is a way to determine the specific
    harmonic at which a waveform reaches \textbf{95\% of its total power}
    during Fourier analysis.

    Unlike the ``trial and error'' method, which relies on manual guessing
    to find the threshold, the algorithm method uses programming structures
    to automate the search, making it less precise in manual effort but
    requiring much less work for the user.

    \subsubsection{\texorpdfstring{\textbf{How the Method
    Works}}{How the Method Works}}\label{how-the-method-works}

    The algorithm follows a logic of iterating through the harmonics and
    checking the accumulated power against a calculated target.

    \begin{enumerate}
        \def\labelenumi{\arabic{enumi}.}
        \tightlist
        \item
        \textbf{Define the Target Power:} First, you must calculate the
        threshold, which is 95\% of the total RMS power. This is represented
        as: \texttt{power95\ =\ 0.95\ *\ (newRMS\^{}2)}.
        \item
        \textbf{Initialize a Counter:} You start with a counter representing
        the first harmonic (e.g., \texttt{harmonic\ =\ 1}).
        \item
        \textbf{Iterative Search (Looping):} You use a \texttt{for} loop to
        cycle through the available harmonics.
        \item
        \textbf{Conditional Check:} Inside the loop, you use an \texttt{if}
        statement to check the current accumulated power using the
        \texttt{sumPowers} toolbox function:

        \begin{itemize}
            \tightlist
            \item
            If the sum of the power of the current harmonics is greater than
            your \texttt{power95} target, the condition is met.
        \end{itemize}
        \item
        \textbf{Store and Terminate:} Once the condition is met, you store
        that specific harmonic number and use the \texttt{break} command to
        exit the loop immediately.
    \end{enumerate}

    \subsubsection{\texorpdfstring{\textbf{Example
    Logic}}{Example Logic}}\label{example-logic}

    The implementation follows this structure:

    \begin{Shaded}
        \begin{Highlighting}[]
            \VariableTok{power95} \OperatorTok{=} \FloatTok{0.95}\OperatorTok{*}\NormalTok{(}\VariableTok{newRMS}\OperatorTok{.\^{}}\FloatTok{2}\NormalTok{)}\OperatorTok{;}
            \VariableTok{harmonic} \OperatorTok{=} \FloatTok{1}\OperatorTok{;}
            \KeywordTok{for} \VariableTok{count}\OperatorTok{=}\FloatTok{1}\OperatorTok{:}\FloatTok{100000}
            \KeywordTok{if} \VariableTok{sumPowers}\NormalTok{(}\VariableTok{dcpt}\OperatorTok{,} \VariableTok{Apt}\NormalTok{(}\FloatTok{1}\OperatorTok{:}\VariableTok{count}\NormalTok{)) }\OperatorTok{\textgreater{}} \VariableTok{power95}
            \VariableTok{harmonic} \OperatorTok{=} \VariableTok{count}\OperatorTok{;}
            \VariableTok{power} \OperatorTok{=} \VariableTok{sumPowers}\NormalTok{(}\VariableTok{dcpt}\OperatorTok{,} \VariableTok{Apt}\NormalTok{(}\FloatTok{1}\OperatorTok{:}\VariableTok{count}\NormalTok{))}\OperatorTok{;}
            \KeywordTok{break}\OperatorTok{;}
            \KeywordTok{end}
            \KeywordTok{end}
        \end{Highlighting}
    \end{Shaded}

    \emph{(Note: While loops and conditional statements are generally
    discouraged in this course, they are explicitly permitted for this
    specific Fourier analysis algorithm )}

    \newpage
    \subsection{Conversation 13: Conceptual Explanation: sumPowers vs. sumFouriers Functions}
    \subsubsection{USER}\label{user-12}

    Explain the difference between sumPowers and sumFouriers

    \subsubsection{ASSISTANT}\label{assistant-12}

    Based on the provided documentation, \texttt{sumPowers} and
    \texttt{sumFouriers} serve two distinct purposes within the Fourier
    analysis workflow: one is used for \textbf{calculating energy/power
    distribution}, while the other is used for \textbf{waveform
    reconstruction}.

    \subsubsection{\texorpdfstring{\textbf{sumPowers}}{sumPowers}}\label{sumpowers}

    The \texttt{sumPowers} function is used to calculate the total power
    contributed by the DC component and the various harmonic components of a
    signal.

    \begin{itemize}
        \tightlist
        \item
        \textbf{Purpose:} It is used to prove that the sum of the waveform's
        individual power components adds up to the total RMS power.
        \item
        \textbf{Logic:} It calculates power based on the assumption of a
        \(1\ \Omega\) resistor, where \(Power = \frac{V_{rms}^2}{1}\).
        \item
        \textbf{Inputs:} It takes the DC voltage (\texttt{dc}) and a vector of
        amplitudes (\texttt{A}) for the harmonics.
        \item
        \textbf{Mathematical Model:} The power is calculated as the sum of the
        DC power and the power of each harmonic:
        \(Power_{total} = \frac{dc^2}{R} + \sum \frac{A^2}{2R}\) (where
        \(R = 1\ \Omega\)).
    \end{itemize}

    \subsubsection{\texorpdfstring{\textbf{sumFouriers}}{sumFouriers}}\label{sumfouriers}

    The \texttt{sumFouriers} function is used to reconstruct the time-domain
    waveform by summing a specific number of Fourier harmonics.

    \begin{itemize}
        \tightlist
        \item
        \textbf{Purpose:} It generates an approximated version of the original
        time-domain waveform using a specified number of harmonics.
        \item
        \textbf{Logic:} It creates a \(t, wv\) vector pair that represents the
        visual shape of the signal after summing the components.
        \item
        \textbf{Inputs:} It requires the number of samples (\texttt{N}), the
        DC component (\texttt{dc}), the frequency vector (\texttt{f}), the
        amplitude vector (\texttt{A}), the phase vector (\texttt{theta}), and
        the specific number of harmonics to sum (\texttt{F}).
        \item
        \textbf{Mathematical Model:} It sums the components in the form
        \(A(n) \cdot \cos(2\pi \cdot n \cdot f(n) \cdot t + \theta(n))\) for
        \(n = 1\) to \(F\).
    \end{itemize}

    \subsubsection{\texorpdfstring{\textbf{Summary
    Comparison}}{Summary Comparison}}\label{summary-comparison}

    {\def\LTcaptype{none} % do not increment counter
        \begin{longtable}[]{@{}
                >{\raggedright\arraybackslash}p{(\linewidth - 4\tabcolsep) * \real{0.3333}}
                >{\raggedright\arraybackslash}p{(\linewidth - 4\tabcolsep) * \real{0.3333}}
                >{\raggedright\arraybackslash}p{(\linewidth - 4\tabcolsep) * \real{0.3333}}@{}}
            \toprule\noalign{}
            \begin{minipage}[b]{\linewidth}\raggedright
                Feature
            \end{minipage} & \begin{minipage}[b]{\linewidth}\raggedright
                \texttt{sumPowers}
            \end{minipage} & \begin{minipage}[b]{\linewidth}\raggedright
                \texttt{sumFouriers}
            \end{minipage} \\
            \midrule\noalign{}
            \endhead
            \bottomrule\noalign{}
            \endlastfoot
            \textbf{Primary Goal} & Quantify power/energy distribution &
            Reconstruct/Approximate the waveform shape \\
            \textbf{Output Type} & A scalar (total power) & A \(t, wv\) vector pair
            (time and waveform) \\
            \textbf{Key Use Case} & Proving power components sum to total RMS power
            & Plotting a waveform using only a subset of harmonics (e.g., 95\%
            power) \\
        \end{longtable}
    }

    \newpage
    \subsection{Conversation 14: Conceptual Explanation: dsolve Function}
    \subsubsection{USER}\label{user-13}

    Explain how to use dsolve

    \subsubsection{ASSISTANT}\label{assistant-13}

    Based on the provided documentation, \texttt{dsolve} is a MATLAB
    function used to solve systems of symbolic differential equations. It
    can handle single differential equations or systems of equations, as
    well as incorporate initial or boundary conditions.

    Here is a breakdown of how to use it:

    \subsubsection{1. Basic Syntax and Setup}\label{basic-syntax-and-setup}

    To use \texttt{dsolve}, you must first define your variables as symbolic
    using the \texttt{syms} command.

    \textbf{Solving a single differential equation:} You define the
    derivative using the \texttt{diff} function. For example, to solve a
    second-order equation:

    \begin{Shaded}
        \begin{Highlighting}[]
            \VariableTok{syms} \VariableTok{y}\NormalTok{(}\VariableTok{t}\NormalTok{)}
            \VariableTok{D2y} \OperatorTok{=} \VariableTok{diff}\NormalTok{(}\VariableTok{y}\OperatorTok{,} \FloatTok{2}\NormalTok{)}\OperatorTok{;} \CommentTok{\% Represents the 2nd derivative of y}
            \VariableTok{Dy} \OperatorTok{=} \VariableTok{diff}\NormalTok{(}\VariableTok{y}\OperatorTok{,} \FloatTok{1}\NormalTok{)}\OperatorTok{;}  \CommentTok{\% Represents the 1st derivative of y}
            \VariableTok{Eqny} \OperatorTok{=} \VariableTok{D2y} \OperatorTok{+} \FloatTok{4}\OperatorTok{*}\VariableTok{Dy} \OperatorTok{+} \FloatTok{3}\OperatorTok{*}\VariableTok{y} \OperatorTok{==} \FloatTok{0}\OperatorTok{;} \CommentTok{\% The symbolic differential equation}
            \VariableTok{y1} \OperatorTok{=} \VariableTok{dsolve}\NormalTok{(}\VariableTok{Eqny}\NormalTok{) }\CommentTok{\% Returns the general solution}
        \end{Highlighting}
    \end{Shaded}

    \textbf{Solving with Initial Conditions (IC):} You can provide specific
    values for the variable at a certain time (e.g., \(t=0\)) to find a
    particular solution.

    \begin{Shaded}
        \begin{Highlighting}[]
            \VariableTok{IC1} \OperatorTok{=} \VariableTok{y}\NormalTok{(}\FloatTok{0}\NormalTok{) }\OperatorTok{==} \FloatTok{5}\OperatorTok{;}
            \VariableTok{IC2} \OperatorTok{=} \VariableTok{Dy}\NormalTok{(}\FloatTok{0}\NormalTok{) }\OperatorTok{==} \FloatTok{1}\OperatorTok{;}
            \VariableTok{IC} \OperatorTok{=}\NormalTok{ [}\VariableTok{IC1} \VariableTok{IC2}\NormalTok{]}\OperatorTok{;} \CommentTok{\% Combine into a vector}
            \VariableTok{y1} \OperatorTok{=} \VariableTok{dsolve}\NormalTok{(}\VariableTok{Eqny}\OperatorTok{,} \VariableTok{IC}\NormalTok{)}
        \end{Highlighting}
    \end{Shaded}

    \subsubsection{2. Complexity of
    Equations}\label{complexity-of-equations}

    \texttt{dsolve} can solve various types of equations: * \textbf{Constant
    forcing functions:} \texttt{Eqny2\ =\ D2y\ +\ 4*Dy\ +\ 3*y\ ==\ 6;} *
    \textbf{Sinusoidal forcing functions:}
    \texttt{Eqny3\ =\ D2y\ +\ 4*Dy\ +\ 3*y\ ==\ 3*cos(3*t);} *
    \textbf{Exponential forcing functions:}
    \texttt{Eqny4\ =\ Dy\ +\ 4*y\ ==\ 4*exp(-2*t);}

    \subsubsection{3. Plotting and Analysis}\label{plotting-and-analysis}

    Once you have obtained a solution with \texttt{dsolve}, you can use
    \texttt{fplot} to visualize the results. This is common when analyzing
    the transient response of a circuit.

    \begin{itemize}
        \tightlist
        \item
        \textbf{Visualizing the solution:} \texttt{fplot(y1,\ {[}0\ 50{]})}
        plots the solution over the time interval from 0 to 50.
        \item
        \textbf{Transient Analysis:} You can use the solutions to calculate
        \textbf{percentage overshoot} or \textbf{percentage undershoot} by
        placing datatips on the resulting plot to find peak and original
        values.
    \end{itemize}

    \subsubsection{4. Application in RLC Reactive
    Circuits}\label{application-in-rlc-reactive-circuits}

    In circuit analysis, \texttt{dsolve} is used to solve for charge (\(q\))
    or current (\(i\)) in RLC circuits. * \textbf{Charge (\(q\)):} You solve
    the equation involving the second derivative of charge
    (\(L \cdot d^2q/dt^2 + R \cdot dq/dt + q/C = \text{forcing function}\)).
    * \textbf{Current (\(i\)):} You can solve for current using an equation
    like \texttt{L*Di\ +\ i*R\ ==\ vS\ -\ q/C}.

    \textbf{Linkage of terms for circuit analysis:} * \textbf{\(L\)} is
    linked with \textbf{inductor} terms. * \textbf{\(R\)} is linked with
    \textbf{resistor} terms. * \textbf{\(C\)} is linked with
    \textbf{capacitor} terms. * \textbf{\(i\)} is linked with
    \textbf{current} terms.

    By solving these equations, you can then use \texttt{fplot} to obtain
    specific voltage information, such as the voltage across the capacitor
    (\(v_C = q/C\)), the voltage across the resistor (\(v_R = i \cdot R\)),
    or the voltage across the inductor (\(v_L\)).

    \newpage
    \subsection{Conversation 15: Extended Theory: Capacitor Voltage with dsolve}
    \subsubsection{USER}\label{user-14}

    Explain how to calculate q/c

    \subsubsection{ASSISTANT}\label{assistant-14}

    In the context of RLC reactive circuit analysis using MATLAB's symbolic
    solver, calculating \(q/C\) (the voltage across a capacitor, \(v_C\)) is
    done by performing a mathematical operation on the symbolic solution for
    charge (\(q\)).

    According to the documentation in \textbf{Module 4: MATLAB symbolic
    solution of CCLODEs}, the relationship between charge and capacitor
    voltage is defined as: \[v_C = \frac{q}{C}\]

    \subsubsection{\texorpdfstring{\textbf{Implementation via
    \texttt{dsolve} and
    \texttt{fplot}}}{Implementation via dsolve and fplot}}\label{implementation-via-dsolve-and-fplot}

    To obtain the voltage profile, you follow these procedural steps:

    \begin{enumerate}
        \def\labelenumi{\arabic{enumi}.}
        \tightlist
        \item
        \textbf{Define the Differential Equation:} Use \texttt{syms\ q(t)} to
        define charge as a symbolic variable and set up the differential
        equation for the circuit (e.g.,
        \(L\frac{d^2q}{dt^2} + R\frac{dq}{dt} + \frac{q}{C} = v_s(t)\)).
        \item
        \textbf{Solve for \(q\):} Use the \texttt{dsolve} function, providing
        the differential equation and any necessary initial conditions (such
        as initial charge \(q(0)\)), to find the symbolic expression for
        \(q\).
        \item
        \textbf{Calculate \(v_C\):} Perform the division of the symbolic
        variable \(q\) by the capacitor constant \(C\).
        \item
        \textbf{Visualize:} Use the \texttt{fplot} function to plot the
        resulting expression \texttt{q/C} over a specified time interval.
    \end{enumerate}

    \textbf{Example Workflow:} As shown in the provided RLC analysis
    example, once you have solved for \(q\) using \texttt{dsolve}, you can
    directly plot the capacitor voltage using the following syntax:

    \begin{Shaded}
        \begin{Highlighting}[]
            \CommentTok{\% After solving for q using dsolve}
            \VariableTok{fplot}\NormalTok{(}\VariableTok{q}\OperatorTok{/}\VariableTok{C}\OperatorTok{,}\NormalTok{ [}\VariableTok{time\_start}\OperatorTok{,} \VariableTok{time\_end}\NormalTok{])}\OperatorTok{;}
        \end{Highlighting}
    \end{Shaded}

    \newpage
    \subsection{Conversation 16: Extended Theory: Inductor Voltage with dsolve}
    \subsubsection{USER}\label{user-15}

    Explain how to calculate the voltage across the inductor

    \subsubsection{ASSISTANT}\label{assistant-15}

    To calculate the voltage across an inductor (\(v_L\)) in a circuit using
    MATLAB, you can use the analytical relationship derived from the
    circuit's differential equations.

    Based on the documentation for \textbf{RLC Reactive Circuit Analysis},
    the voltage across the inductor is found by subtracting the voltage
    across the other components from the source voltage.

    \subsubsection{\texorpdfstring{\textbf{Mathematical
    Relationship}}{Mathematical Relationship}}\label{mathematical-relationship}

    In a series RLC circuit with a source voltage \(v_S\), the sum of the
    voltages across the inductor (\(v_L\)), resistor (\(v_R\)), and
    capacitor (\(v_C\)) must equal the source voltage:
    \[v_S = v_L + v_R + v_C\]

    Therefore, to find the voltage across the inductor, you rearrange the
    equation as: \[v_L = v_S - v_R - v_C\]

    \subsubsection{\texorpdfstring{\textbf{Implementation in
    MATLAB}}{Implementation in MATLAB}}\label{implementation-in-matlab}

    When using \texttt{dsolve} for circuit analysis, you typically follow
    these steps to obtain \(v_L\):

    \begin{enumerate}
        \def\labelenumi{\arabic{enumi}.}
        \tightlist
        \item
        \textbf{Solve for Charge (\(q\)):} Use \texttt{dsolve} to solve the
        differential equation for charge \(q(t)\).
        \item
        \textbf{Calculate Capacitor Voltage (\(v_C\)):} Use the relationship
        \(v_C = q/C\).
        \item
        \textbf{Solve for Current (\(i\)):} Use \texttt{dsolve} to solve for
        the current \(i(t)\).
        \item
        \textbf{Calculate Resistor Voltage (\(v_R\)):} Use the relationship
        \(v_R = i \cdot R\).
        \item
        \textbf{Determine \(v_L\):} Subtract the calculated \(v_R\) and
        \(v_C\) from the source voltage \(v_S\).
    \end{enumerate}

    \textbf{Code Example:} As shown in the RLC Reactive Circuit analysis
    examples, the implementation looks like this:

    \begin{Shaded}
        \begin{Highlighting}[]
            \CommentTok{\% Given the symbolic solutions for q and i from dsolve}
            \VariableTok{vL} \OperatorTok{=} \VariableTok{vP}\OperatorTok{*}\VariableTok{sin}\NormalTok{(}\FloatTok{2}\OperatorTok{*}\VariableTok{pi}\OperatorTok{*}\VariableTok{f}\OperatorTok{*}\VariableTok{t}\NormalTok{) }\OperatorTok{{-}} \VariableTok{q}\OperatorTok{/}\VariableTok{C} \OperatorTok{{-}} \VariableTok{i}\OperatorTok{*}\VariableTok{R}\OperatorTok{;}

            \CommentTok{\% Plot the resulting inductor voltage}
            \VariableTok{fplot}\NormalTok{(}\VariableTok{vL}\OperatorTok{,}\NormalTok{ [}\VariableTok{time\_start}\OperatorTok{,} \VariableTok{time\_end}\NormalTok{])}\OperatorTok{;}
        \end{Highlighting}
    \end{Shaded}

    \newpage
    \subsection{Conversation 17: Extended Theory: Resistor Voltage with dsolve}
    \subsubsection{USER}\label{user-16}

    Explain how to calculate the voltage across the resistor using dsolve

    \subsubsection{ASSISTANT}\label{assistant-16}

    Based on the provided documentation, calculating the voltage across a
    resistor (\(v_R\)) using \texttt{dsolve} is typically part of a larger
    circuit analysis workflow, such as solving for an RLC circuit.

    To calculate \(v_R\), you follow these conceptual steps:

    \subsubsection{\texorpdfstring{1. Solve for the Current
        (\(i\))}{1. Solve for the Current (i)}}\label{solve-for-the-current-i}

    In most circuit models presented in the notes, the voltage across a
    resistor is directly proportional to the current flowing through it
    (\(v_R = i \cdot R\)). Therefore, your first step is to use
    \texttt{dsolve} to find the symbolic expression for the current.

    For an RL circuit, you would define a differential equation for the
    current, such as: \texttt{eqnI\ =\ L*Di\ +\ i*R\ ==\ vS\ -\ q/C;} (where
    \texttt{Di} is the derivative of current) or a similar expression
    depending on the specific circuit topology.

    \subsubsection{2. Apply the Ohm's Law
    Relationship}\label{apply-the-ohms-law-relationship}

    Once \texttt{dsolve} has provided you with the symbolic solution for
    current (\(i\)), you use the relationship \(v_R = i \cdot R\) to derive
    the voltage.

    \textbf{Implementation Logic:} * \textbf{Define the Resistor:} Identify
    the resistance value \(R\). * \textbf{Perform Multiplication:} Multiply
    the symbolic current vector \(i\) by the scalar \(R\). * \textbf{Example
    Syntax:} \texttt{vR\ =\ i\ *\ R;}.

    \subsubsection{3. Visualization}\label{visualization}

    After calculating the expression for \(v_R\), you can use \texttt{fplot}
    to visualize the voltage profile over a specific time interval.

    \textbf{Example Workflow (from RLC Analysis):}

    \begin{Shaded}
        \begin{Highlighting}[]
            \CommentTok{\% 1. Solve for current i using dsolve}
            \VariableTok{eqnI} \OperatorTok{=} \VariableTok{L}\OperatorTok{*}\VariableTok{Di} \OperatorTok{+} \VariableTok{i}\OperatorTok{*}\VariableTok{R} \OperatorTok{==} \VariableTok{vP}\OperatorTok{*}\VariableTok{sin}\NormalTok{(}\FloatTok{2}\OperatorTok{*}\VariableTok{pi}\OperatorTok{*}\VariableTok{f}\OperatorTok{*}\VariableTok{t}\NormalTok{) }\OperatorTok{{-}} \VariableTok{q}\OperatorTok{/}\VariableTok{C}\OperatorTok{;}
            \VariableTok{ICI} \OperatorTok{=} \VariableTok{i}\NormalTok{(}\FloatTok{0}\NormalTok{) }\OperatorTok{==} \FloatTok{0}\OperatorTok{;}
            \VariableTok{i} \OperatorTok{=} \VariableTok{dsolve}\NormalTok{(}\VariableTok{eqnI}\OperatorTok{,} \VariableTok{ICI}\NormalTok{)}\OperatorTok{;}

            \CommentTok{\% 2. Calculate voltage across resistor}
            \VariableTok{vR} \OperatorTok{=} \VariableTok{i} \OperatorTok{*} \VariableTok{R}\OperatorTok{;}

            \CommentTok{\% 3. Plot the result}
            \VariableTok{fplot}\NormalTok{(}\VariableTok{vR}\OperatorTok{,}\NormalTok{ [}\VariableTok{time\_start}\OperatorTok{,} \VariableTok{time\_end}\NormalTok{])}\OperatorTok{;}
        \end{Highlighting}
    \end{Shaded}

    \newpage
    \subsection{Conversation 18: Extended Theory: Series Current with dsolve}
    \subsubsection{USER}\label{user-17}

    Explain how to calculate the current using dsolve

    \subsubsection{ASSISTANT}\label{assistant-17}

    To calculate the current (\(i\)) using \texttt{dsolve}, you must set up
    a symbolic differential equation that describes the relationship between
    the components in your circuit.

    Based on the documentation for \textbf{RLSinForce} and \textbf{RLForce}
    (Module 5), current is typically found by solving the differential
    equation for the circuit's specific topology.

    \subsubsection{1. Set up the Symbolic
    Variables}\label{set-up-the-symbolic-variables}

    First, you must define your time variable and your current variable as
    symbolic.

    \begin{Shaded}
        \begin{Highlighting}[]
            \VariableTok{syms} \VariableTok{i}\NormalTok{(}\VariableTok{t}\NormalTok{)}
        \end{Highlighting}
    \end{Shaded}

    \subsubsection{2. Define the Differential
    Equation}\label{define-the-differential-equation}

    The equation you provide to \texttt{dsolve} depends on the circuit being
    analyzed. The current is determined by the sum of voltages or the
    derivative of charge.

    \begin{itemize}
        \item
        \textbf{For an RL Circuit:} The relationship is defined by the
        inductor (\(L\)), the resistor (\(R\)), and the source voltage
        (\(v_S\)). The differential equation is represented as:
        \[\frac{di}{dt} = \frac{v_S - iR}{L}\] In MATLAB, this is expressed
        as: \texttt{eqnI\ =\ L*Di\ +\ i*R\ ==\ vS;} (where \texttt{Di} is the
        derivative of \texttt{i}).
        \item
        \textbf{For an RLC Circuit:} The current can be derived from the
        charge equation
        (\(L\frac{d^2q}{dt^2} + R\frac{dq}{dt} + \frac{q}{C} = v_s\)). Once
        the symbolic solution for charge \(q\) is found, the current can be
        solved. Specifically, for a series RLC circuit, the current equation
        used in the documentation is:
        \texttt{eqnI\ =\ L*Di\ +\ i*R\ ==\ vP*sin(2*pi*f*t)\ -\ q/C;}.
    \end{itemize}

    \subsubsection{3. Solve with Initial
    Conditions}\label{solve-with-initial-conditions}

    To find a particular solution rather than a general one, you must
    provide initial conditions (e.g., the current at time \(t=0\)). *
    \textbf{Syntax:} \texttt{i\ =\ dsolve(eqnI,\ IC);} where \texttt{IC} is
    a vector containing your initial condition, such as
    \texttt{i(0)\ ==\ 0}.

    \subsubsection{4. Example Workflow}\label{example-workflow}

    According to the \textbf{RLSinForce} example, the calculation follows
    this logic:

    \begin{Shaded}
        \begin{Highlighting}[]
            \CommentTok{\% Define constants}
            \VariableTok{A} \OperatorTok{=} \FloatTok{20}\OperatorTok{;} \VariableTok{f} \OperatorTok{=} \FloatTok{2e03}\OperatorTok{;} \VariableTok{T} \OperatorTok{=} \FloatTok{1}\OperatorTok{/}\VariableTok{f}\OperatorTok{;} \VariableTok{R} \OperatorTok{=} \FloatTok{500}\OperatorTok{;} \VariableTok{L} \OperatorTok{=} \FloatTok{39.8e{-}03}\OperatorTok{;}
            \VariableTok{syms} \VariableTok{i}\NormalTok{(}\VariableTok{t}\NormalTok{)}

            \CommentTok{\% Define the derivative and the differential equation}
            \VariableTok{Di} \OperatorTok{=} \VariableTok{diff}\NormalTok{(}\VariableTok{i}\OperatorTok{,} \FloatTok{1}\NormalTok{)}\OperatorTok{;}
            \VariableTok{vS} \OperatorTok{=} \VariableTok{A}\OperatorTok{*}\VariableTok{sin}\NormalTok{(}\FloatTok{2}\OperatorTok{*}\VariableTok{pi}\OperatorTok{*}\VariableTok{f}\OperatorTok{*}\VariableTok{t}\NormalTok{)}\OperatorTok{;}
            \VariableTok{eqnI} \OperatorTok{=} \VariableTok{L}\OperatorTok{*}\VariableTok{Di} \OperatorTok{+} \VariableTok{i}\OperatorTok{*}\VariableTok{R} \OperatorTok{==} \VariableTok{vS} \OperatorTok{{-}} \VariableTok{q}\OperatorTok{/}\VariableTok{C}\OperatorTok{;} \CommentTok{\% Note: q must be previously solved via dsolve}

            \CommentTok{\% Solve for current with initial condition i(0) = 0}
            \VariableTok{ICI} \OperatorTok{=} \VariableTok{i}\NormalTok{(}\FloatTok{0}\NormalTok{) }\OperatorTok{==} \FloatTok{0}\OperatorTok{;}
            \VariableTok{i} \OperatorTok{=} \VariableTok{dsolve}\NormalTok{(}\VariableTok{eqnI}\OperatorTok{,} \VariableTok{ICI}\NormalTok{)}\OperatorTok{;}
        \end{Highlighting}
    \end{Shaded}

    \newpage
    \subsection{Conversation 19: Conceptual Explanation: Percent Overshoot with dsolve or ode45}
    \subsubsection{USER}\label{user-18}

    Explain the calculation of percent overshoot

    \subsubsection{ASSISTANT}\label{assistant-18}

    The calculation of \textbf{percentage overshoot} is used during the
    analysis of transient responses in differential equations (using
    \texttt{dsolve}) and \texttt{ode45} simulations to determine the nature
    of a system's response.

    According to the documentation, the calculation follows the standard
    formula for percentage change:
    \[\text{percentage overshoot} = \left( \frac{\text{Peak Value} - \text{Original Value}}{\text{Original Value}} \right) \times 100\]

    \subsubsection{\texorpdfstring{\textbf{Step-by-Step
    Procedure}}{Step-by-Step Procedure}}\label{step-by-step-procedure}

    \begin{enumerate}
        \def\labelenumi{\arabic{enumi}.}
        \tightlist
        \item
        \textbf{Identify the ``Original Value'':} In a transient response
        model, the original value is defined as the \textbf{first
        chronological minimum value} at the start of the chart (the baseline
        from which the overshoot occurs).
        \item
        \textbf{Identify the ``Peak Value'':} This is the new maximum value
        reached during the overshoot.
        \item
        \textbf{Apply the Formula:} Subtract the original value from the peak
        value, divide by the original value, and multiply by 100.
    \end{enumerate}

    \subsubsection{\texorpdfstring{\textbf{Practical
    Example}}{Practical Example}}\label{practical-example}

    In the provided RLC circuit analysis documentation, an example is given
    where: * \textbf{Peak overshoot value} = \(3.02\) * \textbf{Original
    value} = \(2.9\)

    The calculation is executed in MATLAB as follows:

    \begin{Shaded}
        \begin{Highlighting}[]
            \VariableTok{percentageOvershoot} \OperatorTok{=}\NormalTok{ ((}\FloatTok{3.02} \OperatorTok{{-}} \FloatTok{2.9}\NormalTok{) }\OperatorTok{/} \FloatTok{2.9}\NormalTok{) }\OperatorTok{*} \FloatTok{100}
        \end{Highlighting}
    \end{Shaded}

    \subsubsection{\texorpdfstring{\textbf{Visualization
    Tip}}{Visualization Tip}}\label{visualization-tip}

    To perform this calculation manually from a generated plot, you should
    use \textbf{datatips} (markers) in the MATLAB figure window. Place one
    marker on the original baseline value and another on the peak overshoot
    value to extract the precise numbers needed for the formula.

    \newpage
    \subsection{Conversation 20: Visualization: Percent Overshoot Chart Markers}
    \subsubsection{USER}\label{user-19}

    Show an example chart for percent overshoot

    \subsubsection{ASSISTANT}\label{assistant-19}

    Based on the provided documentation, an example of a chart used to
    determine percent overshoot is shown below.

    To calculate this, you must place markers (datatips) on the waveform in
    the figure window at the proper points. For the overshoot model, the
    \textbf{original value} is defined as the first chronological minimum
    value at the start of the chart, and the \textbf{peak overshoot value}
    is the new peak reached.

    \begin{figure}
        \centering
        \pandocbounded{\includegraphics[keepaspectratio,alt={Percentage Overshoot}]{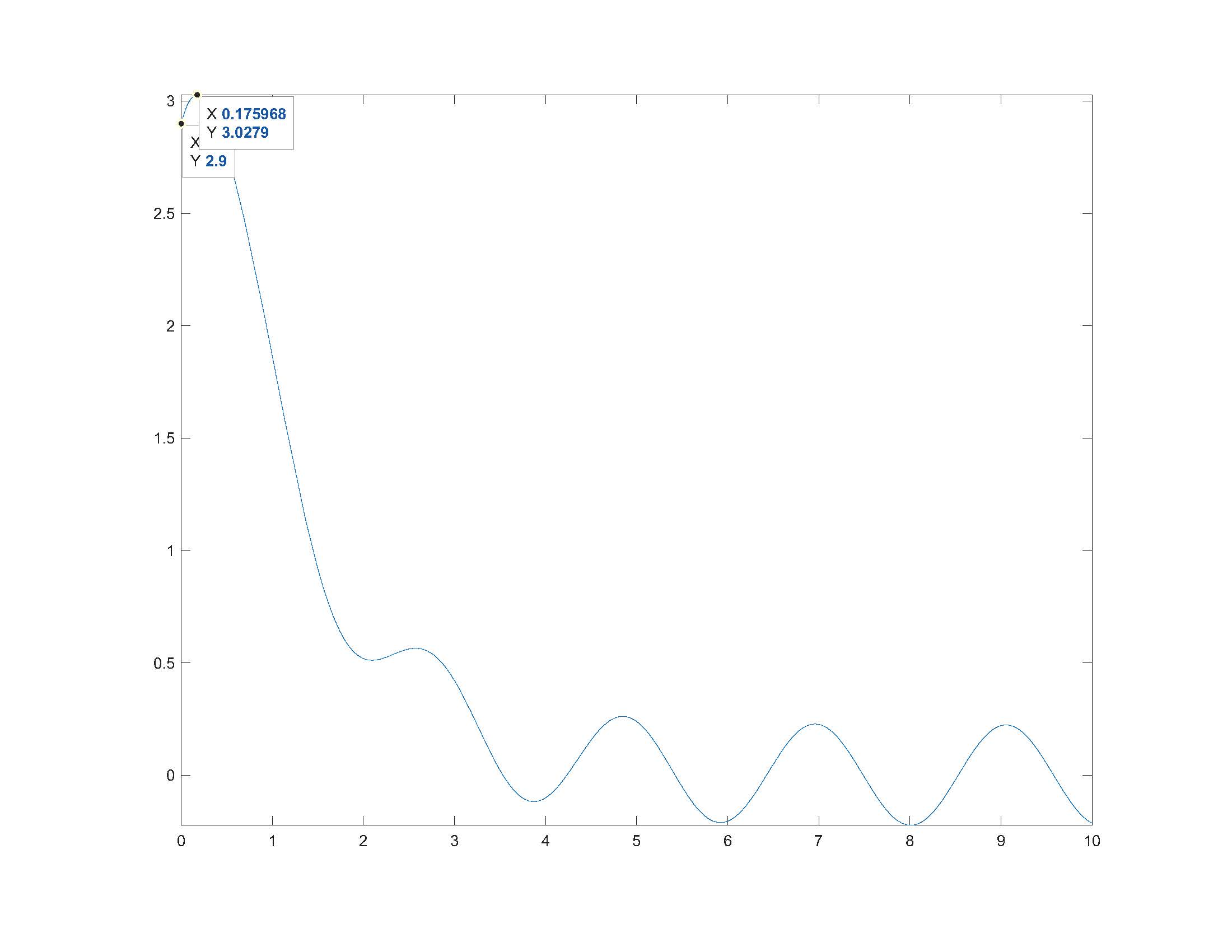}}
        \caption{Percentage Overshoot}
    \end{figure}

    Figure: Example of percentage overshoot chart markers.

    The calculation for the example shown in the documentation is:
    \texttt{percentageOvershoot\ =\ ((3.02\ -\ 2.9)\ /\ 2.9)\ *\ 100}

    \newpage
    \subsection{Conversation 21: Conceptual Explanation: Percent Undershoot with dsolve or ode45}
    \subsubsection{USER}\label{user-20}

    Explain the calculation of percent undershoot

    \subsubsection{ASSISTANT}\label{assistant-20}

    The calculation of \textbf{percentage undershoot} is used to determine
    the magnitude of a downward transient excursion in a waveform. Much like
    the calculation for percent overshoot, it is based on the standard
    percentage change formula:

    \[\text{percentage undershoot} = \left( \frac{\text{Peak Undershoot Value} - \text{Original Value}}{\text{Original Value}} \right) \times 100\]

    According to the course documentation, the specific points used in the
    calculation are defined as follows:

    \begin{itemize}
        \tightlist
        \item
        \textbf{Peak Undershoot Value:} This is the new minimum value reached
        during the undershoot (e.g., a negative excursion).
        \item
        \textbf{Original Value:} This is the second chronological maximum
        (rise) value that occurs directly after the peak undershoot value.
    \end{itemize}

    \subsubsection{\texorpdfstring{\textbf{Example
    Calculation}}{Example Calculation}}\label{example-calculation}

    In the documentation's provided example, if the peak undershoot value is
    \textbf{\(-0.03\)} and the original value is \textbf{\(0.037\)}, the
    calculation would be performed as follows:

    \begin{Shaded}
        \begin{Highlighting}[]
            \VariableTok{percentageUndershoot} \OperatorTok{=}\NormalTok{ ((}\OperatorTok{{-}}\FloatTok{0.03} \OperatorTok{{-}} \FloatTok{0.037}\NormalTok{) }\OperatorTok{/} \FloatTok{0.037}\NormalTok{) }\OperatorTok{*} \FloatTok{100}
        \end{Highlighting}
    \end{Shaded}

    This type of analysis is typically performed during transient response
    studies using \texttt{dsolve} or \texttt{ode45} to characterize how a
    system responds to sudden changes.

    \newpage
    \subsection{Conversation 22: Visualization: Percent Undershoot Chart Markers}
    \subsubsection{USER}\label{user-21}

    Show an example chart for percent undershoot

    \subsubsection{ASSISTANT}\label{assistant-21}

    Based on the provided documentation, an example of a chart used to
    determine percentage undershoot is shown below.

    To calculate this, you must place markers (datatips) on the waveform in
    the figure window at the proper points. For the undershoot model, the
    \textbf{peak undershoot value} is the new minimum value, and the
    \textbf{original value} is the second chronological maximum (rise) value
    located directly after that peak undershoot.

    \begin{figure}
        \centering
        \pandocbounded{\includegraphics[keepaspectratio,alt={Percentage Undershoot}]{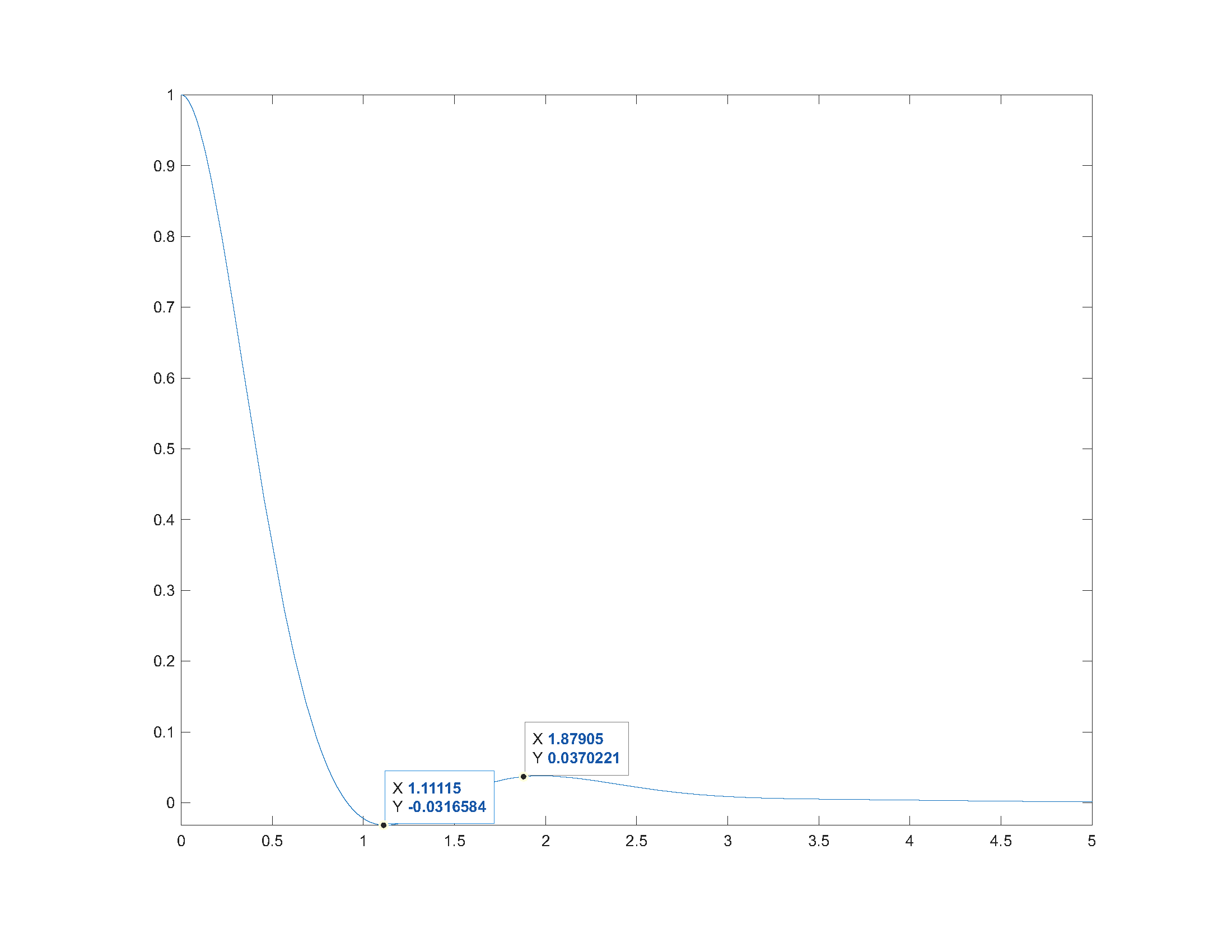}}
        \caption{Percentage Undershoot}
    \end{figure}

    Figure: Example of percentage undershoot chart markers.

    The percentage undershoot calculation for the example in the
    documentation (where the peak undershoot value is \(-0.03\) and the
    original value is \(0.037\)) resolves to:

    \begin{Shaded}
        \begin{Highlighting}[]
            \VariableTok{percentageUndershoot} \OperatorTok{=}\NormalTok{ ((}\OperatorTok{{-}}\FloatTok{0.03} \OperatorTok{{-}} \FloatTok{0.037}\NormalTok{) }\OperatorTok{/} \FloatTok{0.037}\NormalTok{) }\OperatorTok{*} \FloatTok{100}
        \end{Highlighting}
    \end{Shaded}

    \newpage
    \subsection{Conversation 23: Conceptual Explanation: ode45}
    \subsubsection{USER}\label{user-22}

    Explain ode45

    \subsubsection{ASSISTANT}\label{assistant-22}

    In the context of this course, \texttt{ode45} is a MATLAB solver used to
    solve ordinary differential equations (ODEs) numerically. While
    \texttt{dsolve} is used for symbolic solutions, \texttt{ode45} is used
    for numerical simulations, particularly when analyzing the time-domain
    behavior of electrical circuits.

    Based on the documentation for \textbf{Module 5}, here is an explanation
    of how \texttt{ode45} is used:

    \subsubsection{1. Core Purpose and
    Functionality}\label{core-purpose-and-functionality}

    \texttt{ode45} is used to solve differential equations by stepping
    through time. It is frequently applied to analyze the transient
    responses of various circuit types, such as: * \textbf{RL and RC
    circuits} with sinusoidal inputs. * \textbf{RLC circuits} (including
    more complex or high-frequency scenarios). *
    \textbf{Discharging/Charging} behaviors (e.g., a capacitor discharging
    through a resistor).

    \subsubsection{2. The Use of Function
    Handles}\label{the-use-of-function-handles}

    To use \texttt{ode45}, you must provide the differential equation in the
    form of a \textbf{function handle} using the \texttt{@} operator. This
    tells MATLAB how the derivative of your variable (like current \(i\) or
    voltage \(v_C\)) changes with respect to time.

    The solver expects a function that defines the derivative (e.g.,
    \(\frac{di}{dt}\) or \(\frac{dv_C}{dt}\)). An example of a function
    handle for an RL circuit is: \texttt{@(t,\ iL)\ RLForceF(t,\ iL,\ f)}.

    \subsubsection{3. Typical Workflow in Circuit
    Analysis}\label{typical-workflow-in-circuit-analysis}

    The process for using \texttt{ode45} in this course generally follows
    these steps:

    \begin{enumerate}
        \def\labelenumi{\arabic{enumi}.}
        \item
        \textbf{Define the Time Span (\texttt{tSpan}):} Create a vector of
        time points over which you want the solver to calculate the response
        (e.g., \texttt{tSpan\ =\ 0:1e-06:10*T}).
        \item
        \textbf{Define Initial Conditions:} Specify the starting value of the variable at $t=0$ (e.g., \texttt{$i(0) = 0$ or $v_C(0) = 10$}).
        \item
        \textbf{Call the Solver:} Pass the function handle, the time span, and
        the initial conditions to \texttt{ode45}.

        \begin{itemize}
            \tightlist
            \item
            \emph{Syntax:}
            \texttt{{[}t,\ solution{]}\ =\ ode45(function\_handle,\ tSpan,\ initial\_condition);}.
        \end{itemize}
        \item
        \textbf{Analyze the Output:} The solver returns a time vector
        \texttt{t} and the calculated values (like current or voltage) which
        can then be plotted to observe the transient or steady-state response.
    \end{enumerate}

    \subsubsection{4. Practical Applications}\label{practical-applications}

    The course utilizes \texttt{ode45} for several specific analytical
    tasks: * \textbf{Transient Response:} Observing how a system settles
    into a steady state after a change (e.g., an inductor current settling
    after a step change in frequency). * \textbf{Frequency Sweeps:} Running
    the solver at different frequencies (\(f_1, f_2, f_3\), etc.) to observe
    how the system's response (like overshoot or settling time) changes as
    the frequency increases. * \textbf{Determining Settling Time:} Using the
    numerical results to find when a system reaches a steady state or stays
    within a certain percentage (e.g., 5\%) of its final value.

    \newpage
    \subsection{Conversation 24: Conceptual Explanation: ode45 with RC Circuit Models}
    \subsubsection{USER}\label{user-23}

    Explain ode45 with RC

    \subsubsection{ASSISTANT}\label{assistant-23}

    Based on the course documentation, \texttt{ode45} is a numerical solver
    used to solve ordinary differential equations (ODEs) by stepping through
    time. In the context of \textbf{RC (Resistor-Capacitor) circuits}, it is
    used to simulate the transient response of the circuit, such as the
    charging or discharging of a capacitor.

    \subsubsection{\texorpdfstring{\textbf{Core Concept for RC
    Circuits}}{Core Concept for RC Circuits}}\label{core-concept-for-rc-circuits}

    When using \texttt{ode45} for an RC circuit, you are solving for the
    change in voltage across the capacitor (\(v_C\)) over time. The solver
    requires a \textbf{function handle} that defines the derivative of the
    voltage (how the voltage changes) based on the current state of the
    circuit.

    \subsubsection{\texorpdfstring{\textbf{The Workflow for RC
    Simulation}}{The Workflow for RC Simulation}}\label{the-workflow-for-rc-simulation}

    \begin{enumerate}
        \def\labelenumi{\arabic{enumi}.}
        \item
        \textbf{Define the Differential Equation:} For a discharging RC
        circuit, the derivative of the capacitor voltage is defined by the
        relationship: \[\frac{dv_C}{dt} = \frac{-v_C}{RC}\] In the provided
        documentation, this is implemented as a function
        \texttt{RCDischargeF}.
        \item
        \textbf{Establish the Function Handle:} To pass this equation to the
        \texttt{ode45} solver, you must use the \texttt{@} operator to create
        a function handle. This tells the solver to evaluate the derivative at
        each time step.
        \item
        \textbf{Set the Time Span and Initial Conditions:}

        \begin{itemize}
            \tightlist
            \item
            \textbf{\texttt{tSpan}}: You must define a time vector (e.g.,
            \texttt{0:1e-06:70e-03}) that tells the solver the starting time,
            the ending time, and the increment between steps.
            \item
            \textbf{Initial Condition}: You must specify the starting voltage of
            the capacitor at \(t=0\) (e.g., \(10\text{V}\)).
        \end{itemize}
        \item
        \textbf{Execute the Solver:} The syntax for the call is:
        \texttt{{[}t,\ vC{]}\ =\ ode45(@(t,\ vC)\ RCDischargeF(t,\ vC),\ tSpan,\ initial\_voltage);}
    \end{enumerate}

    \subsubsection{\texorpdfstring{\textbf{Example Implementation (RC
    Discharge)}}{Example Implementation (RC Discharge)}}\label{example-implementation-rc-discharge}

    As shown in the \textbf{RCDischarge Example} from Module 5, the process
    looks like this:

    \begin{Shaded}
        \begin{Highlighting}[]
            \CommentTok{\% Define the time span}
            \VariableTok{tSpan} \OperatorTok{=} \FloatTok{0}\OperatorTok{:}\FloatTok{1e{-}06}\OperatorTok{:}\FloatTok{70e{-}03}\OperatorTok{;}

            \CommentTok{\% Solve the ODE for capacitor voltage (vC) }
            \CommentTok{\% Using an initial voltage of 10V}
            \NormalTok{[}\VariableTok{t}\OperatorTok{,} \VariableTok{vC}\NormalTok{] }\OperatorTok{=} \VariableTok{ode45}\NormalTok{(}\OperatorTok{@}\NormalTok{(}\VariableTok{t}\OperatorTok{,} \VariableTok{vC}\NormalTok{) }\VariableTok{RCDischargeF}\NormalTok{(}\VariableTok{t}\OperatorTok{,} \VariableTok{vC}\NormalTok{)}\OperatorTok{,} \VariableTok{tSpan}\OperatorTok{,} \FloatTok{10}\NormalTok{)}\OperatorTok{;}

            \CommentTok{\% Plot the result}
            \VariableTok{plot}\NormalTok{(}\VariableTok{t}\OperatorTok{/}\FloatTok{1e{-}03}\OperatorTok{,} \VariableTok{vC}\NormalTok{)}\OperatorTok{;}
        \end{Highlighting}
    \end{Shaded}

    \subsubsection{\texorpdfstring{\textbf{Analyzing the
    Result}}{Analyzing the Result}}\label{analyzing-the-result}

    Once the solver completes, you can use the resulting \texttt{t} and
    \texttt{vC} vector pairs to: * \textbf{Visualize the decay}: Plot the
    voltage vs.~time to see how the capacitor discharges. * \textbf{Time
    Constant Analysis}: You can analyze the curve in relation to the time
    constant (\(\tau = RC\)). For example, the documentation notes that it
    typically takes roughly 5 time constants to reach steady state (0V).

    \newpage
    \subsection{Conversation 25: Study Guide: Midterm Exam}
    \subsubsection{USER}\label{user-25}

    What do I have to study for the midterm exam?

    \subsubsection{ASSISTANT}\label{assistant-25}

    Based on the course documentation, your midterm exam covers
    \textbf{Modules 0 through 3}. To prepare effectively, you should focus
    your studies on the following core topics and procedural workflows:

    \subsubsection{1. MATLAB Fundamentals \& Vectorization (Module
    0)}\label{matlab-fundamentals-vectorization-module-0}

    \begin{itemize}
        \tightlist
        \item
        \textbf{Smart Indexing:} Master the use of index vectors and the
        \texttt{find} function to extract specific data points or ranges.
        Remember that MATLAB indexing is 1-based.
        \item
        \textbf{Vectorization:} Understand the difference between
        \textbf{vector-smart functions} (like \texttt{sin}, \texttt{exp},
        \texttt{sqrt}) that apply operations element-wise and
        \textbf{scalar-aggregating functions} (like \texttt{sum},
        \texttt{mean}, \texttt{max}, \texttt{min}) that collapse vectors into
        single values.
        \item
        \textbf{Conversion Divisors:} Know how to use conversion divisors to
        scale time or voltage/current vectors for engineering notation (e.g.,
        dividing by \texttt{1e-03} to display in milliseconds).
        \item
        \textbf{The \texttt{ones} Function:} Understand how to create constant
        vectors using the syntax \texttt{K\ *\ ones(1,\ n)}.
    \end{itemize}

    \subsubsection{\texorpdfstring{2. Boolean Vectors and the \texttt{find}
    Function (Module
    1)}{2. Boolean Vectors and the find Function (Module 1)}}\label{boolean-vectors-and-the-find-function-module-1}

    \begin{itemize}
        \tightlist
        \item
        \textbf{Boolean Criteria:} Be able to define a criterion using
        relational operators (e.g., \texttt{==}, \texttt{\textgreater{}},
        \texttt{\textless{}=}) and logical operators (e.g., \texttt{\&},
        \texttt{\textbar{}}) to compare a matrix or vector.
        \item
        \textbf{The \texttt{find} Function:} Practice using \texttt{find} to
        return the indices of elements that satisfy a specific Boolean
        criterion.
        \item
        \textbf{Statistical Parsing:} Familiarize yourself with using
        statistical functions like \texttt{mean()}, \texttt{median()},
        \texttt{mode()}, \texttt{sort()}, \texttt{min()}, and \texttt{max()}
        to analyze data.
    \end{itemize}

    \subsubsection{3. Piecewise Continuous Functions (Module
    2)}\label{piecewise-continuous-functions-module-2-1}

    You should master the \textbf{Single Period Method} to construct complex
    waveforms. This involves: * \textbf{Step A: Definition.} Creating
    individual time and function vectors using toolbox functions such as
    \texttt{makeRamp}, \texttt{makeSinewave}, \texttt{expRise}, or
    \texttt{expDecay}. All segments must use the same time increment. *
    \textbf{Step B: Concatenation.} * \textbf{Time Vectors:} Using the
    offset method (\texttt{t\ =\ {[}t\ t(end)\ +\ t2(2:end){]}}) to join
    time segments without duplicating the transition point. *
    \textbf{Function Vectors:} Concatenating values in a single line while
    skipping the first element of subsequent vectors to match the time
    vector's dimensions (\texttt{f\ =\ {[}f1\ f2(2:end){]}}). * \textbf{Step
    C: Periodicity.} Using the \texttt{makeCycles} function to replicate
    your single-period ``template'' into multiple cycles. * \textbf{Special
    Shapes:} Understand how to build square waves by adjusting the duty
    cycle of the high/low segments.

    \subsubsection{4. Fourier Analysis (Module
    3)}\label{fourier-analysis-module-3-1}

    This is a significant portion of the material. Focus on the following
    workflow: * \textbf{Normalization:} Using \texttt{get\_rms\_value} to
    normalize waveforms to a specified RMS voltage. * \textbf{Data
    Extraction:} Using the \texttt{fouriersA} toolbox function to extract
    the DC component, frequencies, amplitudes, and phase angles. *
    \textbf{Power Analysis:} * Proving that the sum of harmonic powers
    equals the total RMS power (where power is calculated across a
    \(1\ \Omega\) resistor, \(P = V_{rms}^2/1\)). * Using \texttt{sumPowers}
    to calculate these sums. * \textbf{Waveform Reconstruction:} Using
    \texttt{sumFouriers} to approximate a waveform from its Fourier
    components. * \textbf{95\% Power Threshold:} Knowing how to determine
    the number of harmonics required to reach 95\% of total power using
    either the \textbf{trial and error method} or the \textbf{algorithm
    method} (which uses loops and conditionals). * \textbf{Phase vs.~Power:}
    Understanding that a \textbf{time shift} in a waveform changes the phase
    components (\(\theta\)) but leaves the power spectrum (amplitudes)
    unchanged.

    \newpage
    \subsection{Conversation 26: Study Guide: Final Exam}
    \subsubsection{USER}\label{user-26}

    What do I have to study for the final exam?

    \subsubsection{ASSISTANT}\label{assistant-26}

    Based on the course documentation, the final exam covers \textbf{all
    modules} (Modules 0 through 5). To prepare, you should master the
    following topics and procedural workflows:

    \subsubsection{1. MATLAB Fundamentals \& Vectorization (Module 0 \&
    1)}\label{matlab-fundamentals-vectorization-module-0-1}

    \begin{itemize}
        \tightlist
        \item
        \textbf{Smart Indexing \& Vectorization:} Master using index vectors
        and the \texttt{find} function to extract data from \texttt{t,\ v,\ i}
        vector pairs. You must understand the difference between
        \textbf{vector-smart functions} (e.g., \texttt{sin}, \texttt{exp},
        \texttt{sqrt}) and \textbf{scalar-aggregating functions} (e.g.,
        \texttt{sum}, \texttt{mean}, \texttt{max}, \texttt{min}).
        \item
        \textbf{Conversion Divisors:} Know how to scale time or
        voltage/current vectors for engineering notation (e.g., dividing by
        \texttt{1e-03} for milliseconds).
        \item
        \textbf{Boolean Logic:} Be proficient in creating Boolean criteria
        using relational operators (e.g., \texttt{==},
        \texttt{\textgreater{}}, \texttt{\textless{}=}) and logical operators
        (e.g., \texttt{\&}, \texttt{\textbar{}}) to be used with the
        \texttt{find} function.
        \item
        \textbf{Statistical Parsing:} Understand how to use functions like
        \texttt{mean()}, \texttt{median()}, \texttt{mode()}, \texttt{sort()},
        \texttt{min()}, and \texttt{max()} to analyze datasets.
    \end{itemize}

    \subsubsection{2. Piecewise Continuous Functions (Module
    2)}\label{piecewise-continuous-functions-module-2-2}

    You must be able to construct complex periodic waveforms using the
    \textbf{Single Period Method}: * \textbf{Step A: Definition.} Use
    toolbox functions like \texttt{makeRamp}, \texttt{makeSinewave},
    \texttt{expRise}, or \texttt{expDecay} to define segments. All segments
    must share the same time increment. * \textbf{Step B: Concatenation.} *
    \textbf{Time Vectors:} Use the offset method
    (\texttt{t\ =\ {[}t\ t(end)\ +\ t2(2:end){]}}) to join segments without
    duplicating the transition point. * \textbf{Value Vectors:} Concatenate
    function values in a single line, skipping the first element of
    subsequent vectors to ensure dimension matching
    (\texttt{f\ =\ {[}f1\ f2(2:end){]}}). * \textbf{Step C: Periodicity.}
    Use \texttt{makeCycles} to replicate a single-period template into a
    multi-cycle waveform. * \textbf{Special Shapes:} Learn to build square
    waves by managing the duty cycle of high/low segments.

    \subsubsection{3. Fourier Analysis (Module
    3)}\label{fourier-analysis-module-3-2}

    This module focuses on analyzing the frequency components of signals: *
    \textbf{Normalization:} Use \texttt{get\_rms\_value} to normalize
    waveforms to a specific RMS voltage. * \textbf{Data Extraction:} Use
    \texttt{fouriersA} to obtain the DC component, frequencies, amplitudes,
    and phase angles. * \textbf{Power Analysis:} * Understand that power is
    calculated across a \(1\ \Omega\) resistor (\(P = V_{rms}^2/1\)). * Use
    \texttt{sumPowers} to verify that the sum of harmonic powers equals the
    total RMS power. * \textbf{Waveform Reconstruction:} Use
    \texttt{sumFouriers} to approximate a waveform using a specific number
    of harmonics. * \textbf{95\% Power Threshold:} Be able to determine how
    many harmonics are required to reach 95\% of total power using either
    the \textbf{trial and error method} or the \textbf{algorithm method}
    (using loops and conditionals). * \textbf{Phase vs.~Power:} Understand
    that a \textbf{time shift} changes the phase components (\(\theta\)) but
    leaves the power spectrum (amplitudes) unchanged.

    \subsubsection{4. Symbolic Solutions \& CCLODEs (Module
    4)}\label{symbolic-solutions-cclodes-module-4}

    \begin{itemize}
        \tightlist
        \item
        \textbf{Symbolic Solving:} Use \texttt{dsolve} to solve differential
        equations for circuits.
        \item
        \textbf{Circuit Linkage:} Link \textbf{\(L\)} with inductor terms,
        \textbf{\(R\)} with resistor terms, \textbf{\(C\)} with capacitor
        terms, and \textbf{\(i\)} with current terms.
        \item
        \textbf{Transient Analysis:} Use \texttt{fplot} to visualize solutions
        and perform transient response analysis, such as calculating
        \textbf{percentage overshoot} and \textbf{undershoot}.
    \end{itemize}

    \subsubsection{\texorpdfstring{5. Function Handles \& \texttt{ode45}
    Solver (Module
    5)}{5. Function Handles \& ode45 Solver (Module 5)}}\label{function-handles-ode45-solver-module-5}

    \begin{itemize}
        \tightlist
        \item
        \textbf{Function Handles:} Understand the \texttt{@} operator used to
        pass equations to solvers.
        \item
        \textbf{Numerical Simulation:} Use the \texttt{ode45} solver to
        simulate the time-domain behavior of circuits, including:

        \begin{itemize}
            \tightlist
            \item
            \textbf{RL and RC circuits} with sinusoidal inputs.
            \item
            \textbf{RLC circuits}.
        \end{itemize}
        \item
        \textbf{Transient Behavior:} Use the solver to analyze settling times,
        charging/discharging, and frequency response (e.g., observing how
        overshoot changes with frequency).
    \end{itemize}

\end{document}